# How Well Do Large-Scale Chemical Language Models Transfer to Downstream Tasks?

**Tatsuya Sagawa**[1,2] and **Ryosuke Kojima**[2,3]
[1] Graduate School of Pharmaceutical Sciences, Kyoto University
[2] RIKEN BDR
[3] Graduate School of Medicine, Kyoto University
sagawa.tatsuya.82s@st.kyoto-u.ac.jp, kojima.ryosuke.8e@kyoto-u.ac.jp

## Abstract

Chemical Language Models (CLMs) pre-trained on large-scale molecular data are widely used for molecular property prediction. However, the common belief that increasing training resources, such as model size, dataset size, and training compute, improves both pre-training loss and downstream task performance has not been systematically validated in the chemical domain. In this work, we evaluate this assumption by pre-training CLMs while scaling training resources and measuring transfer performance across diverse molecular property prediction (MPP) tasks. We find that while pre-training loss consistently decreases with increased training resources, downstream task performance shows limited improvement. Moreover, alternative metrics based on the Hessian or loss landscape also fail to estimate downstream performance in CLMs. We further identify conditions under which downstream performance saturates or degrades despite continued improvements in pre-training metrics, and analyze the underlying task-dependent failure modes through parameter space visualizations. These results expose a gap between pre-training-based evaluation and downstream performance, and emphasize the need for model selection and evaluation strategies that explicitly account for downstream task characteristics. The code is available at https://github.com/sagawatatsuya/MolScaleTransfer

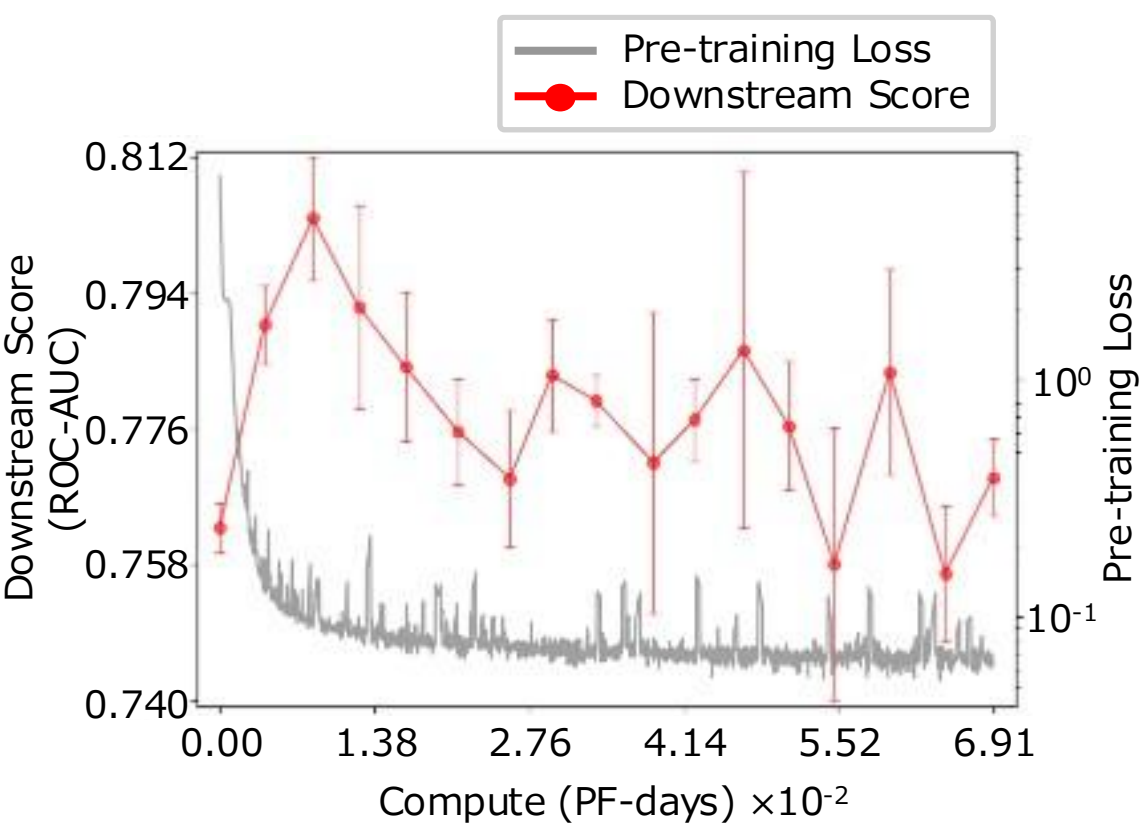

Figure 1: Training dynamics of the pre-training cross-entropy loss in a CLM and the downstream performance (ROC-AUC) obtained by fine-tuning checkpoints from different pre-training stages on the HIV benchmark.

## 1 Introduction

Chemical language models (CLMs), which are pre-trained on string representations of molecules, have become increasingly important as foundation models that can transfer to a wide range of molecular property prediction (MPP) tasks in biology, chemistry, and drug discovery [Park et al., 2024; Ross et al., 2022; Edwards et al., 2022]. This is largely because labeled data are expensive to obtain, while the rapid growth of molecular databases has made billions of unlabeled structures available. By representing molecules as sequences of atomic and bond symbols, we can treat them as a form of language. This formulation makes it natural to adopt techniques from natural language processing (NLP), and a variety of NLP methods have been adopted in this line of work [Chithrananda et al., 2020; Li and Jiang, 2021; Irwin et al., 2022]. Motivated by scaling practices in NLP, CLMs have also been rapidly scaled, following the expectation that increasing model size, data size, and compute lead to better performance [Soares et al., 2025a, 2025b; Cai et al., 2025]. In NLP, pre-training loss has been shown to follow a power law with respect to training resources, and this relationship has served as a practical guideline for model design and training resource allocation [Kaplan et al., 2020; Hoffmann et al., 2022].

In recent years, studies in NLP have reported that improvements in pre-training loss do not necessarily translate into better downstream performance and can even lead to negative transfer in some cases [Zoph et al., 2020; Isik et al., 2024; Lourie et al., 2025]. This observation suggests that the implicit assumption that minimizing the pre-training loss is equivalent to acquiring useful representations may not be universal. To address this issue, prior work has proposed alternatives to pre-training loss, including Hessian and loss landscape-based measures, and has shown that these metrics can serve as indicators of transfer performance on downstream NLP tasks [Liu et al., 2023]. However, whether negative transfers occur and whether such alternative metrics to pre-training loss are expected to be effective depend on the downstream tasks and application domains. In particular, these questions have not been sufficiently examined in the context of CLMs.

Figure 1 illustrates that, for a typical Transformer-based CLM, lower pre-training loss does not necessarily translate into better downstream performance. As pre-training progresses, the loss decreases monotonically, but downstream performance after fine-tuning model states from different stages on the HIV benchmark [Wu et al., 2018] is non-monotonic. This suggests that optimizing the pre-training objective can diverge from learning representations that transfer well. It also indicates that common CLM practices, such as using pre-training loss as the primary criterion for early stopping, model selection, and training resource allocation, may not be fully justified from the perspective of downstream applications.

Motivated by this observation, we investigate (i) how well pre-training loss correlates with downstream performance on MPP tasks, and (ii) how this relationship changes as we scale model size, data size, and training compute. To address these questions, we focus on encoder-based CLMs that are commonly used for MPP, pre-train them under different resource budgets and evaluate transfer on MPP benchmarks. We then evaluate downstream performance under practical transfer settings that are widely used in CLM applications, including fine-tuning and linear probe, to identify when improvements in pre-training align with downstream gains versus when they diverge. We further analyze task-dependent factors underlying this phenomenon in a comprehensive evaluation across 36 MPP tasks.

Our contributions are summarized as follows:

- We show that scaling model size, data size, and training compute consistently reduces pre-training loss, suggesting scaling-law behavior for CLMs at least with respect to loss.
- We systematically evaluate the relationship between pre-training loss and downstream performance on MPP benchmarks and demonstrate that lower loss does not reliably imply better downstream performance in CLMs.
- We show that, under standard CLM training and transfer settings, the relationship between pre-training loss and downstream performance depends strongly on the task and the training setup, which establishes the limitations of using pre-training loss as a single criterion for early stopping, model selection, and resource allocation. These limitations persist even when using alternatives to pre-training loss based on Hessian information and loss landscapes. Finally, we analyze this behavior using visualizations in parameter space.

# 2 Related Work

## 2.1 Chemical Language Models for Molecular Property Prediction

Chemical language models (CLMs) are typically pre-trained on large-scale unlabeled corpora of molecular strings and use representations such as SMILES (Simplified Molecular-Input Line-Entry System) as input [Weininger, 1988; Park et al., 2024; Ross et al., 2022; Yüksel et al., 2023]. This allows molecules to be treated as text. Accordingly, CLMs are commonly pre-trained with standard NLP objectives, including masked language modeling (MLM), which predicts masked tokens, and autoregressive (next-token) language modeling. The same framework is widely used for downstream molecular property prediction (MPP) tasks, covering a broad range of classification and regression problems.

Input representations for CLMs have been studied extensively. Beyond SMILES, alternative notations such as Self-Referencing Embedded Strings (SELFIES) [Krenn et al., 2020] and DeepSMILES [O'Boyle and Dalke, 2018] have been developed and have also been used in CLM pre-training and modeling [Yüksel et al., 2023]. However, prior work reports that the choice of notation has little impact on downstream performance for large-scale CLMs [Rajan et al., 2026]. In light of this observation, our scaling study focuses on SMILES, which remains widely used in large-scale CLMs.

## 2.2 Scaling Laws in Chemical Language Models

Scaling laws have been observed across domains such as NLP, computer vision, and speech recognition, where performance improves predictably as model size, data size, and training compute increase [Zhai et al., 2021; Chen et al., 2025; Kaplan et al., 2020]. Similar observations have been reported in the chemistry domain. For example, prior work shows that pre-training loss in decoder-based CLMs follows a power law with respect to model size and data size [Ramos et al., 2025; Frey et al., 2023]. For encoder-based models, which are often pre-trained and then fine-tuned for MPP tasks, existing studies have typically examined pre-training or downstream performance in isolation [Chen et al., 2023]. As a result, how strongly improvements in pre-training loss translate into downstream gains remains unclear.

Scaling-law studies also extend beyond string-based CLMs. In graph-based molecular representation learning, performance has been reported to scale consistently with downstream dataset size [Chen et al., 2023]. Another line of work studies models that incorporate atomic coordinates and reports monotonic improvements on certain quantum chemistry benchmarks as model size increases [Wang et al., 2024]. While these approaches go beyond simple string representations of molecules, combining their insights with our comprehensive downstream-task results on standard string representation may help clarify the mechanisms underlying these scaling phenomena.

## 2.3 When Pre-training Loss Fails to Reflect Transferability

In research on pre-trained models, the empirical rule that lower pre-training loss implies better downstream performance has guided practical choices such as model selection, early stopping, and resource allocation [Kaplan et al., 2020]. Evidence from large language models (LLMs) has further reinforced this practice. As scale increases, loss often decreases alongside improvements in downstream performance, which encourages the use of loss as a proxy for downstream performance.

However, recent studies have shown that improvements in pre-training loss do not necessarily yield downstream gains and can even lead to negative transfer [Zoph et al., 2020; He et al., 2019]. For example, under substantial distribution shift between pre-training and downstream data, downstream performance may not scale predictably with increasing amounts of pre-training data [Isik et al., 2024]. In addition, a meta-analysis of LLMs found that the conditions under which downstream scaling laws hold are limited and that scaling behavior can fail to generalize even under small changes in experimental settings [Lourie et al., 2025]. These findings suggest that making pre-training-stage decisions based solely on loss can be unreliable in some settings. To better anticipate downstream performance beyond what is captured by loss, prior work has proposed alternatives to pre-training loss, such as metrics derived from the loss-landscape curvature [Liu et al., 2023] and gradient-based measures that quantify similarity between parameter vectors before and after learning [Yao et al., 2024].

Negative transfer has also been reported in molecular modeling. Prior work on CLMs shows that, in certain tasks and data regimes, pre-training contributes little to downstream performance and can even degrade transfer [Chen et al., 2023; Hormazabal et al., 2024]. Studies of graph-based molecular representations likewise report that pre-training can induce negative transfer, highlighting the importance of pre-training task design and data conditions [Xia et al., 2022]. In such cases, improvements in pre-training loss do not necessarily indicate that the model has learned representations useful for downstream tasks. Nevertheless, pre-training loss remains a widely used metric for monitoring pre-training progress and selecting models. This may be partly because much of the literature examines pre-training and downstream performance in isolation and reports results in relatively specific settings. As a result, comprehensive evaluations that systematically vary model size, data size, and compute under a consistent pre-training protocol are still limited. Moreover, alternatives to pre-training loss that have shown promise in NLP have not yet been systematically examined for CLMs.

# 3 Method

We systematically vary model size, data size, and training compute under a controlled pre-training protocol to examine scaling behavior and downstream transfer. We first describe the pre-training and downstream evaluation procedures, and then define metrics quantifying downstream transfer performance.

## 3.1 Pre-training and Downstream Evaluation

**Pre-training**. We pre-train CLMs with MLM on a large-scale corpus of SMILES strings. We randomly mask a subset of tokens and train the model to predict the masked tokens. We adopt a Transformer encoder architecture [Vaswani et al., 2017] (see Appendix A for details). The pre-training objective is the MLM cross-entropy loss, which we track throughout training.

**Downstream tasks**. We evaluate downstream transfer using two standard protocols: fine-tuning and linear probe. For fine-tuning, we attach a task-specific prediction head to the pre-trained encoder and update both the encoder and head to optimize the downstream objective. This setting reflects performance after task-specific adaptation of the pre-trained representations. For linear probe, we freeze the encoder and train only a linear head on top of the encoder representations, measuring the linear separability of the learned representations. For each benchmark, we use task-specific losses and evaluation metrics (Appendix B).

## 3.2 Downstream Transferability

We quantitatively evaluate how well pre-training loss correlates with downstream performance. We compare model states saved at different stages of pre-training and test whether pre-training loss exhibits a monotonic relationship with downstream metrics via correlation analysis.

We periodically save the model during pre-training and refer to each saved snapshot as a checkpoint.

For each checkpoint, we compute the following values:

- $L_{\text{pre}}$: the MLM cross-entropy loss on the validation split of the pre-training dataset
- $L_{\text{down}}$: the MLM cross-entropy loss computed on downstream inputs without any parameter updates (i.e. zero-shot evaluation)
- $P^{\text{FT}}$: downstream performance after fine-tuning
- $P^{\text{LP}}$: downstream performance under linear probe

We define consistency as the monotonic association between $L_{\text{pre}}$ and downstream metrics and quantify it using Spearman's rank correlation coefficient [Spearman, 1904], denoted by $\rho$. Specifically, we compute $\rho$ between $L_{\text{pre}}$ and each of $L_{\text{down}}$, $P^{\text{FT}}$, and $P^{\text{LP}}$ across checkpoints. To keep the direction consistent across tasks, we transform downstream metrics only for correlation computation so that smaller values indicate better performance; for classification tasks, we use $1 - \text{ROC-AUC}$.

## 3.3 Alternative Metrics to Pre-training Loss

In addition to the loss-based analysis in Section 3.2, we consider two alternatives to pre-training loss as predictors of downstream performance: the trace of the Hessian ($\text{Tr}(\boldsymbol{H})$) [Liu et al., 2023] and the Principal Gradient-based Measurement (PGM) distance [Yao et al., 2024]. $\text{Tr}(\boldsymbol{H})$ captures the curvature of the pre-training loss landscape, while the PGM distance assesses transferability via the misalignment between gradients induced by the pre-training objective and those induced by downstream objective. We expect smaller values of these metrics to correspond to better downstream performance. Definitions and implementation details are provided in Appendix C.

For each training run with fixed model size and training data size, we compute these metrics on checkpoints saved at regular intervals. We quantify the relationship between metric values and downstream performance using Spearman's rank correlation.

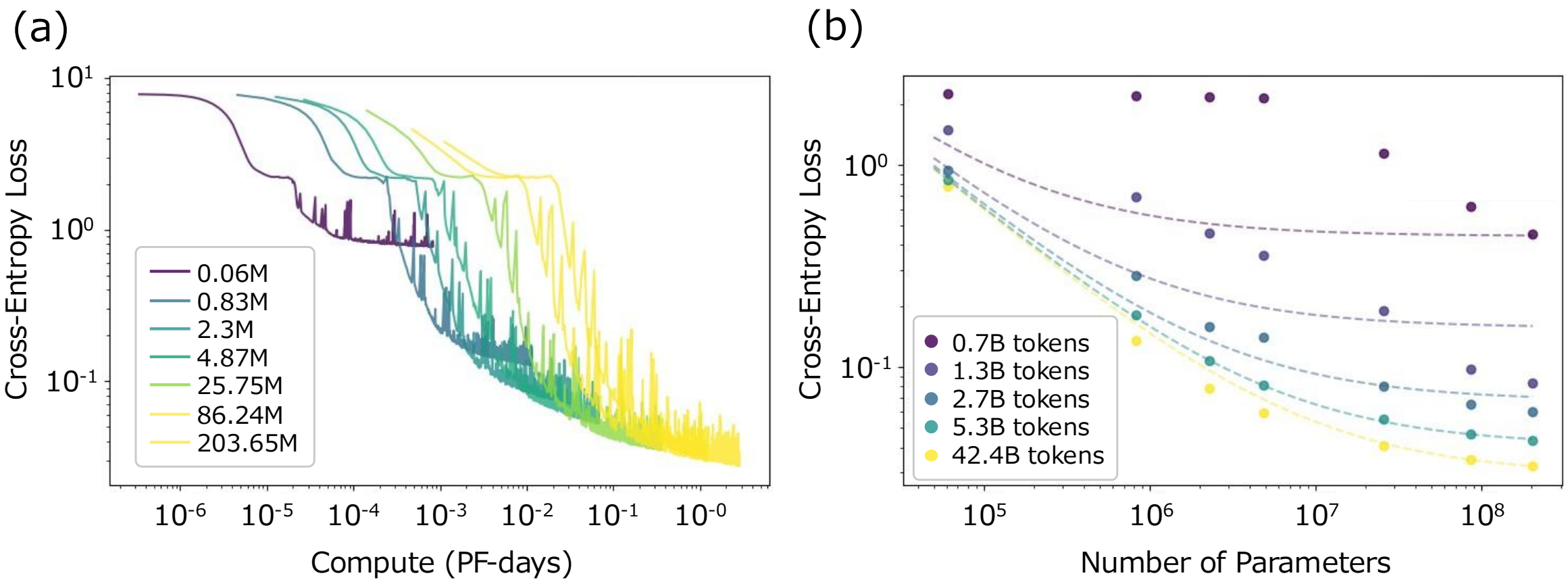

Figure 2: Pre-training scaling laws in pre-training. (a) Pre-training loss as a function of training compute for different model sizes. (b) Final pre-training loss as a function of model size and data size, with a power-law fit (exponents $\alpha = 0.686, \beta = 1.702$).

## 4 Experiments

We examine how scaling affects pre-training loss and downstream transfer performance. Section 4.1 confirms that previously reported pre-training scaling laws hold in our setting. Section 4.2 evaluates models pre-trained under different scaling regimes on downstream tasks and quantifies the relationship between pre-training loss and downstream performance using rank correlation and identifies when loss improvements translate into downstream gains. Finally, Section 4.3 evaluates alternatives to pre-training loss in settings where loss alone fails to predict downstream outcomes and analyzes task-dependent failure modes when using loss as a proxy for transfer.

### 4.1 Pre-training Evaluation

We evaluate how pre-training loss changes as we scale model size, data size, and compute. Our models are Transformer encoders with from 0.06 to 203.65 million parameters excluding embeddings (Appendix A), covering a range comparable to prior encoder-based CLMs [Ross et al., 2022; Park et al., 2024; Yüksel et al., 2023]. We pre-train on a large-scale dataset of SMILES strings compiled from ZINC-15 [Sterling and Irwin, 2015] and PubChem [Kim et al., 2019]. To assess the effect of data size, we subsample the training data to obtain subsets ranging from 0.7 to 42.4 billion tokens.

Figure 2(a) shows learning curves for seven model sizes trained for four epochs with the data size fixed at the largest subset. For all models, pre-training loss decreases throughout training with diminishing returns later in training, and larger models consistently achieve lower loss.

Figure 2(b) shows the final pre-training loss while varying both model size and training data size. Pre-training loss decreases monotonically as model size and data size increase, consistent with scaling-law behavior over our experimental range (Appendix D).

### 4.2 Downstream Transferability Evaluation

We evaluate how improvements during pre-training relate to downstream gains along three axes: model size, data size, and compute. For each axis, we vary only that factor while holding the others fixed. We then plot $L_{\text{pre}}$, $L_{\text{down}}$, $P^{\text{FT}}$, and $P^{\text{LP}}$ across the resulting checkpoints (Figure 3) and quantify the monotonic relationship between $L_{\text{pre}}$ and downstream metrics using Spearman's rank correlation (Figure 4).

For downstream evaluation, we use MoleculeNet [Wu et al., 2018], a widely used benchmark suite for MPP. We evaluate 36 tasks in total: seven classification tasks, five non-quantum regression tasks, and 24 quantum-chemistry regression tasks. Unless otherwise noted, $P^{\text{FT}}$ and $P^{\text{LP}}$ denote performance averaged over each task group. We use ROC-AUC for classification, MAE for quantum-chemistry regression tasks, and RMSE for non-quantum regression tasks (see Appendix B for details).

To disentangle the contributions of training resources to downstream performance, we conduct three scaling experiments. In the model-size experiment, we fix the training data size at 42.4 billion tokens and train for four epochs, comparing seven models with different numbers of parameters. In the data-size experiment, we fix the model at 4.87 million parameters and train for four epochs across five data-size settings ranging from 0.7 to 42.4 billion tokens. In the compute experiment, we fix the model (4.87M parameters) and training data size (42.4 billion tokens) and save checkpoints every 0.25 epoch over the 4-epoch training and compare 17 checkpoints in total, including the 0-step initialization. We measure compute in PF-days [Kaplan et al., 2020; Appendix E].

Figure 3 shows that $L_{\text{pre}}$ decreases consistently as we scale model size, data size, or compute. We also observe the same monotonic decrease for $L_{\text{down}}$, the MLM loss computed on downstream inputs. In other words, despite distributional differences between pre-training and downstream data, improvements under the pre-training objective carry over to downstream inputs, consistent with prior findings [Frey et al., 2023; Isik et al., 2024].

In contrast, downstream performance shows qualitatively different trends depending on which factor is scaled. Scaling model size tends to improve both $P^{\text{FT}}$ and $P^{\text{LP}}$ (Figure 3(a)– (c)). When scaling data size, increasing the number of training tokens from 0.7B to 2.7B typically improves downstream performance; however, further scaling to 5.3B led to diminishing returns and performance degradation (Figure 3(d)–(f)).

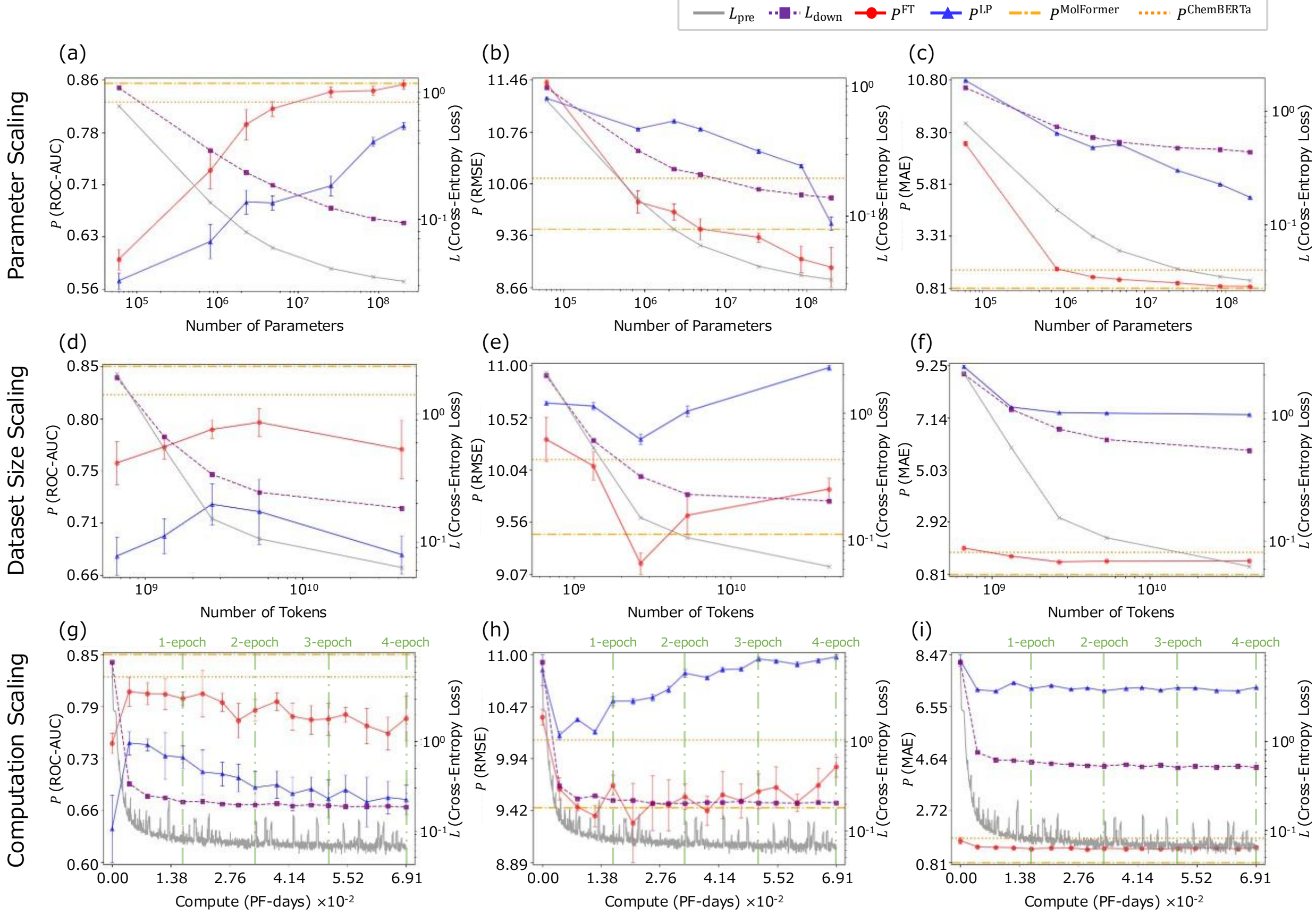

Figure 3: Trends of $L_{\mathrm{pre}}$, $L_{\mathrm{down}}$, $P^{\mathrm{FT}}$, and $P^{\mathrm{LP}}$ under (a–c) model size scaling, (d–f) data size scaling, and (g–i) compute scaling. Panels correspond to task groups: (a,d,g) classification, (b,e,h) non-quantum regression, and (c,f,i) quantum-chemistry regression (averaged over tasks). In (g–i), dashed vertical lines mark epoch boundaries during pre-training. For reference, we include fine-tuned performance of MolFormer-XL ($P^{\mathrm{MolFormer}}$) [Ross et al., 2022] and ChemBERTa ($P^{\mathrm{ChemBERTa}}$) [Chithrananda et al., 2020].

Comparing $P^{\mathrm{FT}}$ and $P^{\mathrm{LP}}$, the benefits of scaling training resources are generally more pronounced under fine-tuning. In particular, as compute increases, $P^{\mathrm{FT}}$ saturates early, while $P^{\mathrm{LP}}$ tends to deteriorate as pre-training proceeds (Figure 3(g)–(i)). In this setting, downstream performance reaches its peaks before completing one epoch ($1.73 \times 10^{-2}$ PF-days), i.e., before the model completes a full pass through the training data. Although CLMs are often pre-trained for multiple epochs [Ross et al., 2022; Frey et al., 2023], these results suggest that longer pre-training can further reduce pre-training loss while potentially causing over-optimizing for the pre-training objective, leading to worse transfer.

As shown in Figure 4, we compute Spearman's rank correlation between $L_{\mathrm{pre}}$ and downstream metrics $P^{\mathrm{FT}}$ and $P^{\mathrm{LP}}$ for each task, as described in Section 3.2, after transforming metrics, when necessary, so that smaller values indicate better performance. The spread and sign of the correlations depend on which factor is scaled. When scaling model size, points lie mostly in the first quadrant, indicating that $\rho(L_{\mathrm{pre}}, P^{\mathrm{FT}})$ and $\rho(L_{\mathrm{pre}}, P^{\mathrm{LP}})$ are typically positive. When scaling data size, points are concentrated in the first quadrant, but several also appear in the second and fourth quadrants. When scaling compute, points are more broadly dispersed, and 26 out of 36 tasks have negative $\rho(L_{\mathrm{pre}}, P^{\mathrm{LP}})$.

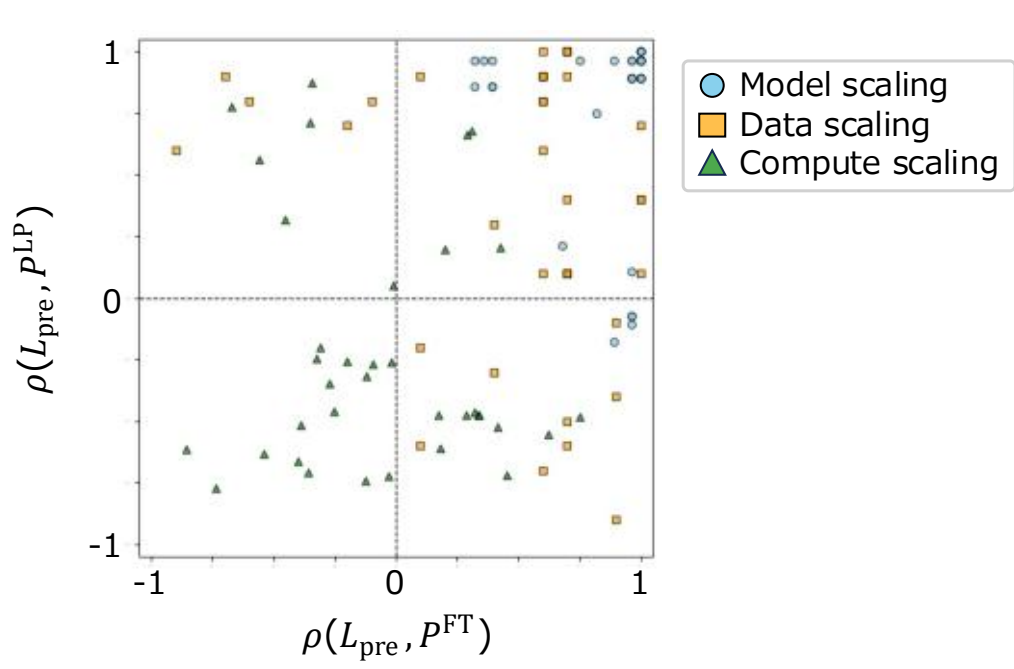

Figure 4: Consistency evaluation using Spearman's rank correlation. When scaling model size, data size, or compute in isolation, the x-axis shows $\rho(L_{\mathrm{pre}}, P^{\mathrm{FT}})$ and the y-axis shows $\rho(L_{\mathrm{pre}}, P^{\mathrm{LP}})$. Each point corresponds to a task.

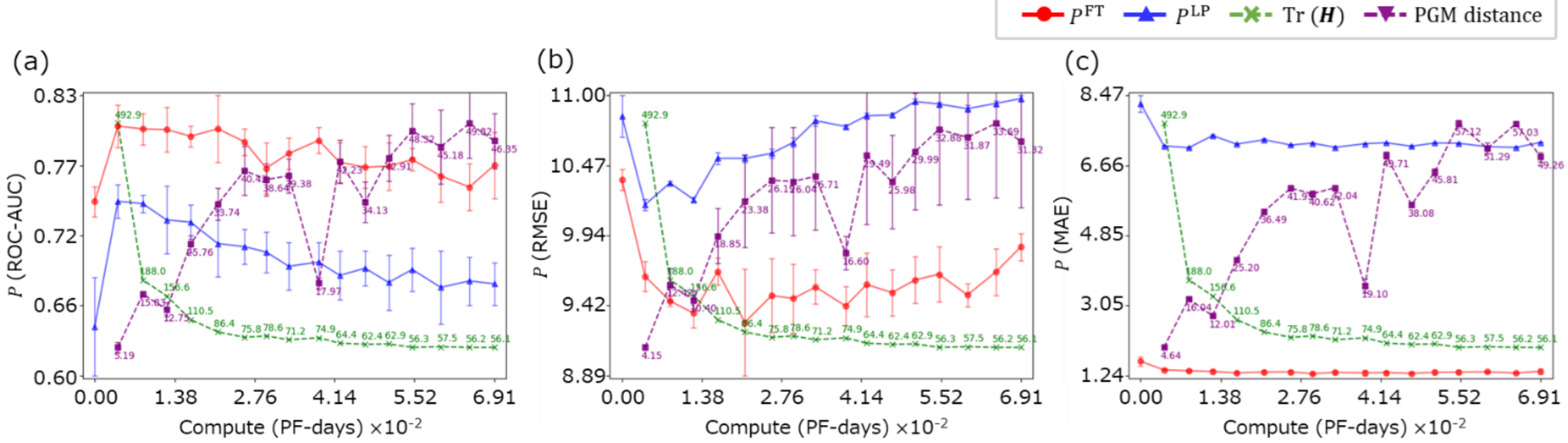

Figure 5: Trajectories of metric values and downstream performance as a function of pre-training compute. The x-axis indicates pre-training progress measured in compute. Subplots correspond to task groups: (a) classification, (b) non-quantum regression, and (c) quantum-chemistry regression. For each group, we plot the task-averaged downstream performance under fine-tuning and linear probe ($P^{\mathrm{FT}}$ and $P^{\mathrm{LP}}$) and overlay the task-averaged PGM distance. We also overlay $\mathrm{Tr}(\boldsymbol{H})$, which is task-agnostic and thus shared across task groups.

Combined with the results in Section 4.1, these findings show that, under commonly used CLM training setups, pre-training loss decreases as we scale model size, data size, or compute, whereas downstream performance varies substantially across tasks and scaling settings. This implies that pre-training loss alone is an unreliable indicator for consistently predicting downstream gains. More broadly, our results suggest that common CLM practices such as early stopping and checkpoint selection based solely on pre-training loss, as well as resource allocation guided by loss-based scaling laws, may fail to deliver consistent downstream improvements.

## 4.3 Transferability Evaluation Using Alternatives to Pre-training Loss

Following the evaluation protocol in Section 3.3, we examine how two metrics beyond pre-training loss, $\mathrm{Tr}(\boldsymbol{H})$ and the PGM distance, relate to downstream metrics ($P^{\mathrm{FT}}$ and $P^{\mathrm{LP}}$). We compute these metrics on checkpoints saved every 0.25 epoch during pre-training and report their trajectories alongside downstream performance in Figure 5 (see Appendix I for per-task results). Note that $\mathrm{Tr}(\boldsymbol{H})$ is task-agnostic and yields a single value per checkpoint. As a result, when evaluated on the same checkpoint sequence, it is not expected to capture task-specific differences in downstream performance trajectories. We further report per-task Spearman correlations between each metric and downstream performance (Figure 6).

In Figure 5, $\mathrm{Tr}(\boldsymbol{H})$ decreases approximately monotonically as pre-training proceeds, whereas downstream performance shows saturation and fluctuations, highlighting intervals where the two curves are not well aligned. This is reflected in Figure 6, where the Spearman correlation between $\mathrm{Tr}(\boldsymbol{H})$ and $P^{\mathrm{FT}}$ varies from negative to positive across tasks. For $P^{\mathrm{LP}}$, we also observe negative correlations, which may indicate that $\mathrm{Tr}(\boldsymbol{H})$ does not capture the late-stage decline in $P^{\mathrm{LP}}$.

Taken together, these results indicate that although $\mathrm{Tr}(\boldsymbol{H})$ and the PGM distance change as pre-training proceeds, neither provides a reliable predictor of downstream performance. $\mathrm{Tr}(\boldsymbol{H})$ primarily reflects progress in optimizing the pre-training objective and decreases in tandem with $L_{\mathrm{pre}}$ and $L_{\mathrm{down}}$.

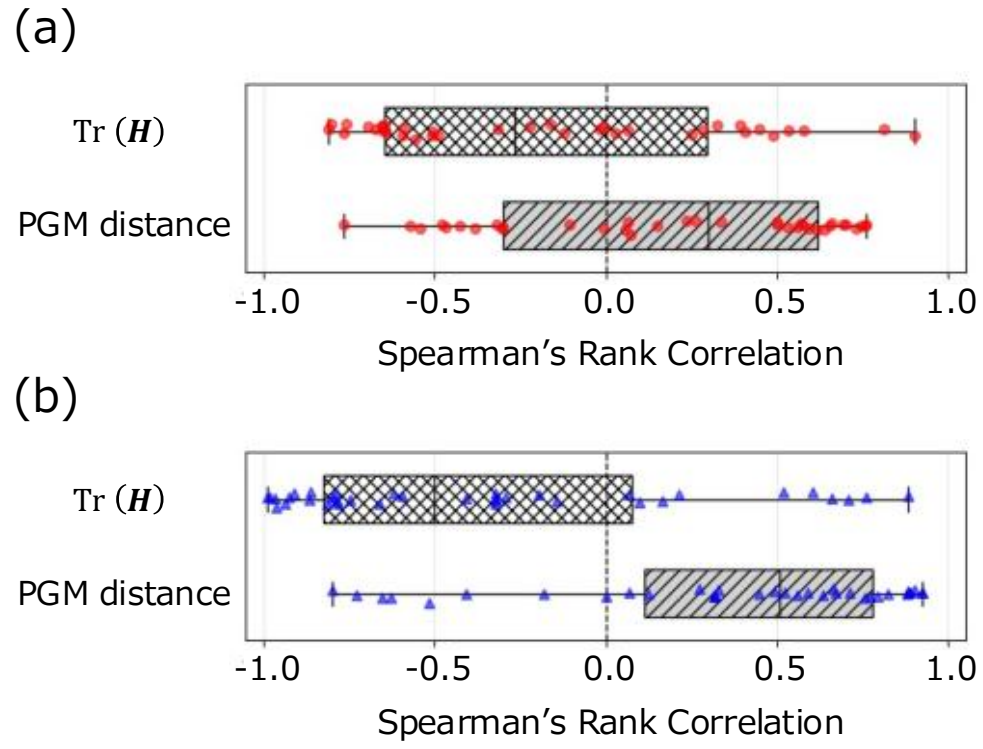

Figure 6: Per-task relationships between metrics computed during pre-training and downstream performance. Spearman correlations between $\mathrm{Tr}(\boldsymbol{H})$ and the PGM distance and (a) $P^{\mathrm{FT}}$ and (b) $P^{\mathrm{LP}}$.

However, its correspondence to task-dependent performance metrics such as $P^{\mathrm{FT}}$ and $P^{\mathrm{LP}}$ is limited. In contrast, the PGM distance shows a weak correlation with $P^{\mathrm{LP}}$, while its trajectory is similar across tasks, suggesting limited sensitivity to task-specific variation in downstream performance.

Overall, our results suggest that relying on a single metric beyond pre-training loss may be insufficient for predicting downstream performance or selecting checkpoints. Accurately estimating downstream performance may instead require lightweight task-aware evaluation, such as periodically evaluating candidate checkpoints or scaling configurations on a small validation set from the target downstream task. Although such heuristic workarounds by task-aware evaluation require access to target-task validation data in advance, this approach offers a practical way to select model size, dataset size, and compute budget.

## 4.4 Visualization of Parameter Space

In this section, we visualize the parameter space to gain insights into why the PGM distance does not reliably predict $P^{\mathrm{FT}}$ and why the consistency between pre-training loss and downstream performance varies by task. We consider 16 checkpoints saved at 0.25-epoch intervals over a 4-epoch pre-training run (excluding the 0-step checkpoint). For each checkpoint, we fine-tune separate models for each of the 36 downstream tasks, resulting in task-specific models. For both the pre-trained checkpoints and the fine-tuned models, we extract the parameters from the final Transformer block and project them into two dimensions using principal component analysis (PCA) (Figure 7(a)). Hereafter, we refer to the pre-training checkpoint used to initialize fine-tuning as the initialization checkpoint. We focus on the final Transformer block because layers closer to the prediction head tend to exhibit more task-specific changes [Mosbach et al., 2020]. We further focus on the checkpoint after 4 epochs and the corresponding fine-tuned models to examine the parameter changes in more detail (Figure 7(b)).

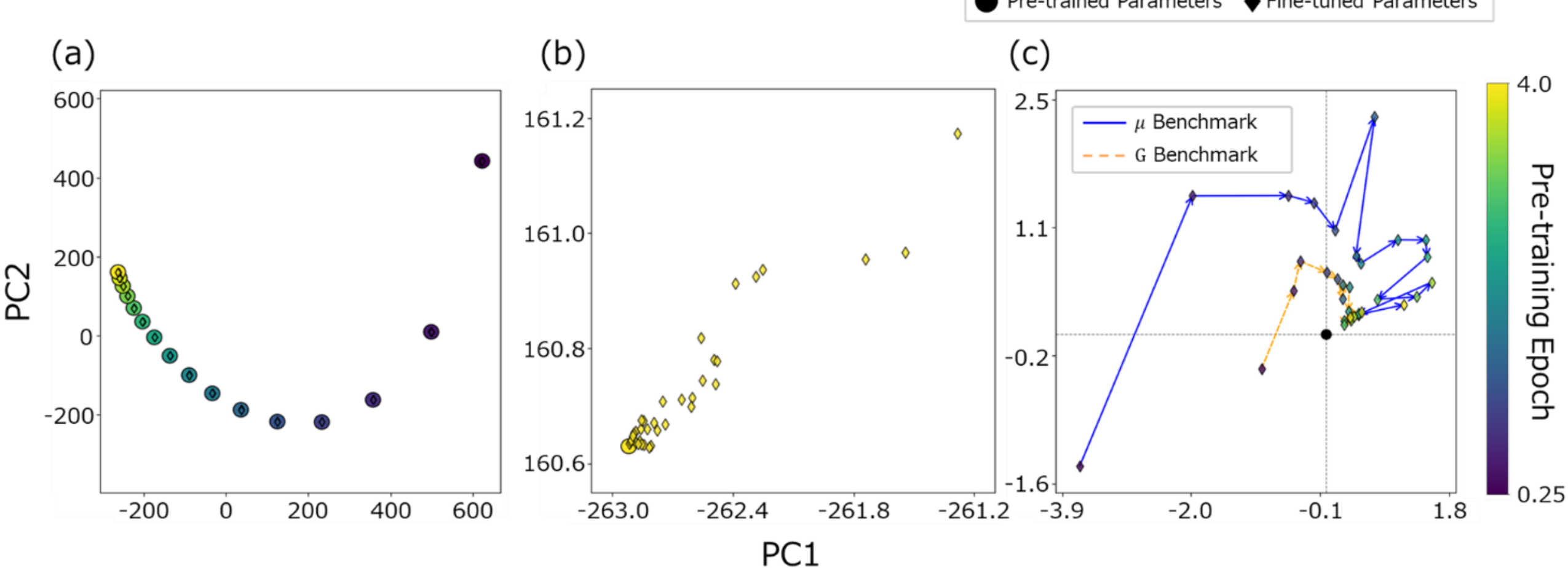

Figure 7: Visualization of the parameter space. (a) Sixteen pre-training checkpoints and the models obtained by fine-tuning each checkpoint separately on each of the 36 downstream tasks are plotted in the same 2D PCA space. (b) A zoomed-in view focusing on the checkpoint after 4 epochs of pre-training and the corresponding fine-tuned models across tasks. (c) Visualization in relative coordinates, obtained by translating each fine-tuned model's point in the projected space so that its initialization checkpoint is mapped to the origin.

Figures 7(a) and 7(b) show that the fine-tuned models remain close to their initialization checkpoints in the projected space. The PGM distance is motivated by the assumption that the pre-training update direction is indicative of a global direction toward a good downstream optimum. However, the visualization suggests that fine-tuning updates remain largely local to each initialization checkpoint, and that the fine-tuned solutions vary substantially across initializations and tasks. As a result, the assumption underlying the PGM distance may not hold in our setting, which makes it difficult to predict downstream performance consistently using a single metric.

Next, to understand the task dependence of the correspondence between loss and downstream performance, we visualize changes induced by fine-tuning as displacements from the initialization checkpoint. As in Figures 7(a) and 7(b), we use the parameters of the final Transformer block and first project the corresponding parameter vectors into a two-dimensional space using PCA. We then compute the displacement of each fine-tuned model from its corresponding initialization checkpoint by translating coordinates so that the initialization checkpoint is mapped at the origin (Figure 7(c)). As a concrete example, we compare two QM9 tasks, $G$ and $\mu$, which exhibit high and low correlations ($\rho = 0.78$ and $\rho = 0.07$), respectively.

As shown in Figure 7(c), $G$ exhibits smaller displacements than $\mu$, indicating that the fine-tuned models remain closer to their initialization checkpoints in the projected space. Although the fine-tuned models for both tasks show decreasing displacement as pre-training proceeds, $G$ remains closer to its initialization checkpoint than $\mu$ at the final stage. We observe qualitatively similar trends across other downstream tasks, although the magnitude varies by task (see Appendix J for details).

## 5 Conclusions

In this study, we systematically examined whether the conventional wisdom that scaling training resources improves performance also holds for downstream transfer performance in chemical language models (CLMs). By independently varying model size, data size, and compute within practical ranges and evaluating on multiple molecular property prediction (MPP) tasks, we found that although pre-training loss consistently decreases as training resources increase, these loss reductions do not reliably translate into downstream gains; in particular, scaling data size and compute often yields limited benefits and can even induce negative transfer under certain conditions. We further showed that neither pre-training loss nor alternative proxy metrics reliably predict downstream performance, implying that common CLM practices such as early stopping, model selection, and compute budgeting based solely on pre-training loss may be suboptimal for downstream transfer. Overall, our findings highlight the need to rethink evaluation metrics and development guidelines for scaling CLMs as foundation models with downstream transfer in mind, and suggest that task-aware evaluation that accounts for task-specffific characteristics of downstream chemical tasks is essential for more efficient and practical foundation-model design.

## Acknowledgments

This work was supported by JST Moonshot R&D (No. JPMJMS2024), JSPS KAKENHI (Nos. JP21H05207, JP21H05221), RIKEN TRIP initiative (AGIS) and CREST (No. JPMJCR22D3).

| $d_{model}$ | $n_{head}$ | $n_{layer}$ | $d_{ff}$ | #parameters(M) |
|---|---|---|---|---|
| 64 | 1 | 1 | 256 | 0.06 |
| 128 | 2 | 4 | 512 | 0.83 |
| 192 | 3 | 5 | 768 | 2.30 |
| 256 | 4 | 6 | 1,024 | 4.87 |
| 512 | 8 | 8 | 2,048 | 25.75 |
| 768 | 12 | 12 | 3,072 | 86.24 |
| 1,024 | 16 | 16 | 4,096 | 203.65 |

Table 1: Model configurations used in this study (number of layers, embedding dimension, attention heads, and FFN intermediate size). "#parameters (M)" denotes the number of non-embedding parameters.

# A Model Design and Pre-training Setup

To examine how well pre-training loss predicts downstream performance, we train chemical language models (CLMs) under a controlled pre-training protocol while systematically varying model size and data size. In the following, we describe the CLM architecture, the construction of pre-training data, and implementation details.

## A.1 Model Architecture

Our CLM is a Transformer encoder that takes SMILES sequences as input. Model size is specified by the embedding dimension, the number of attention heads, the number of layers, and the intermediate dimension of the feed-forward network (FFN). The exact configurations used at each scale are summarized in Table 1. We report parameter counts excluding embeddings. Implementation details follow prior work [Izsak et al., 2021; Liu et al., 2023].

## A.2 Pre-training Data

We construct the pre-training corpus from molecular data in ZINC-15 [Sterling and Irwin, 2015] and PubChem [Kim et al., 2019] (Table 2). Each molecule is converted to a SMILES string, and we perform canonicalization using RDKit[1] (version 2025.3.5). We then remove molecules that RDKit cannot parse, duplicates, and molecules whose tokenized sequence length exceeds 512. Finally, we randomly hold out 10% of the compounds to form a validation set and use the remaining compounds for training. Once created, this validation set is fixed and shared across all training conditions.

**Tokenizer**. Following prior work, we use an atom-wise tokenizer [Ross et al., 2022; Park et al., 2024]. This tokenizer covers atoms, SMILES-specific symbols, ions, and isotopes, and uses a vocabulary of 2362 tokens. We set the maximum input length to 512 tokens. Note that the vocab size reported in Table 2 denotes the number of unique tokens that appear in each dataset.

**Data scaling**. To evaluate the effect of data size, we subsample from the full training corpus (42.4 billion tokens) and create five training datasets of size 0.7, 1.3, 2.7, 5.3, and 42.4 billion tokens. To ensure that differences across scales are not driven by sampling variance, we use nested sampling so that each smaller dataset is a subset of the next larger dataset. In addition, we keep the validation set identical across all conditions and evaluate loss on the same validation set irrespective of the training data size.

## A.3 Pre-training Objective

We pre-train the model using masked language modeling (MLM). For each input sequence, we randomly select 15% of tokens for prediction: 80% of the selected tokens are replaced with a mask token, 10% are replaced with a random token, and the remaining 10% are left unchanged. The model is trained to reconstruct the selected tokens, and the loss is defined as the mean cross-entropy over the selected (masked) positions [Devlin et al., 2019].

## A.4 Optimization and Training Details

We use AdamW [Loshchilov and Hutter, 2017] with $\beta_1 = 0.9$ and $\beta_2 = 0.98$. We linearly warm up the learning rate for 5,000 steps and then keep it at $1 \times 10^{-3}$. We apply weight decay of 0.01 and dropout of 0.1. The batch size is 4,096. We train for 4 epochs and save checkpoints every 0.25 epoch. Training is performed using bfloat16 precision.

All experiments are conducted using a single NVIDIA H200 GPU. Training for 4 epochs on the 42.4B-token dataset corresponds to 95k optimization steps. The wall-clock training time for models with 0.06/0.83/2.3/4.87/25.75/86.24/203.65 million parameters are 4/16/31/36/91/228/481 hours, respectively.

## A.5 Reporting Pre-training Loss

We report pre-training loss evaluated on a held-out validation set from the pre-training dataset. For each checkpoint, we compute the validation loss and present learning curves for each model size and data size setting in Figure 8.

# B Downstream Task Setup

| Dataset | #samples | #tokens | Token length (min/mean/max) | Vocab size |
|---|---|---|---|---|
| ZINC-15 | 997,598,730 | 42,347,563,545 | (4, 43.08, 152) | 113 |
| PubChem | 111,378,206 | 4,793,849,259 | (1, 43.24, 2,211) | 2,349 |

Table 2: Statistics of the pre-training data. For each dataset, we report the number of samples, the total number of tokens, sequence length, and vocabulary size. Token length is defined as the sequence length produced by the atom-wise tokenizer.

[1] https://www.rdkit.org/

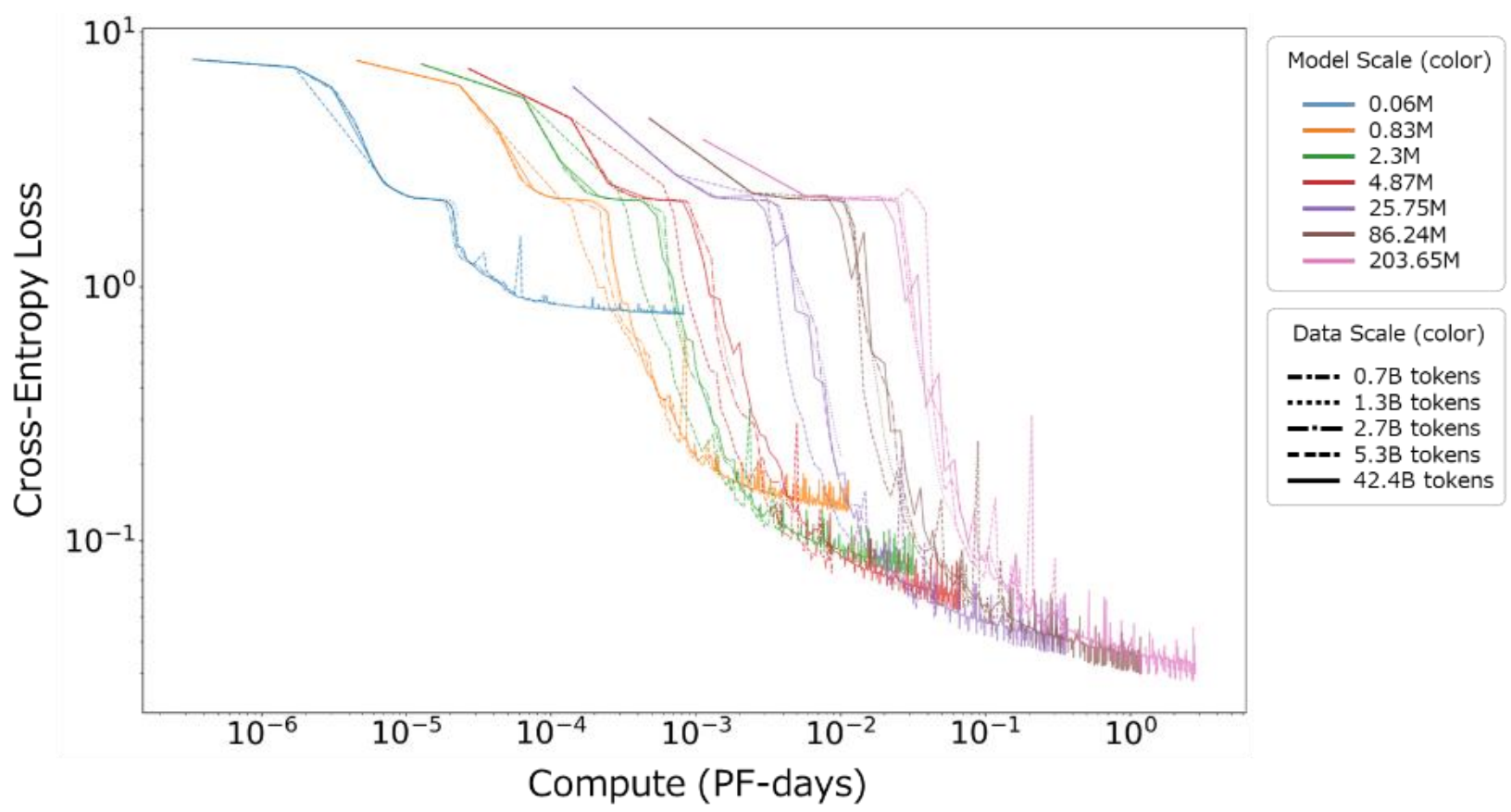

Figure 8: Pre-training loss curves across checkpoints for each model-size and data-size setting. Loss is computed on a held-out validation set.

For downstream evaluation, we quantify transfer performance of pre-trained CLMs on molecular property prediction (MPP) benchmarks using MoleculeNet [Wu et al., 2018], a widely used suite for MPP. For each benchmark, we train and evaluate models using the official train/validation/test splits provided by MoleculeNet. The splitting strategy and dataset statistics for each benchmark are summarized in Table 3.

## B.1 Benchmarks, Splits, and Evaluation Metrics

We use ROC-AUC for classification tasks. For regression tasks, we use MAE for the quantum-chemistry benchmarks QM8 and QM9, and RMSE for the other regression tasks. For all tasks, we run training with three random seeds and report the mean and standard deviation of test-set performance.

For regression tasks, we normalize target values based on the training split so that the targets have mean 0 and standard deviation 1 during training. At evaluation time, we inverse-transform the predictions back to the original scale and then compute MAE or RMSE.

In the three-seed setting, following MoleculeNet's recommendation, we keep each benchmark's split fixed and vary only training stochasticity. Final test evaluation is performed using the checkpoint selected by early stopping based on validation loss.

## B.2 Molecular Representations and Prediction Heads

For each downstream task, we feed the SMILES sequence into the pre-trained encoder and construct a molecular representation from the final-layer hidden states. We then input this representation into a linear prediction head to predict task-specific molecular properties. Input tokenization and the maximum sequence length follow the settings in Appendix A.

Following [Ross et al., 2022], we evaluate two transfer settings: fine-tuning and linear probe. In both settings, molecular representation is computed as the mean over token embeddings from the final layer, excluding special tokens such as <bos>, <eos>, and <pad>. In fine-tuning, we train both the encoder and the prediction head, whereas in linear probe, we freeze the encoder and train only the prediction head.

## B.3 Downstream Objectives

Downstream tasks consist of single-task classification and regression, and multi-task classification. For classification, we use cross-entropy loss; for regression, we use the mean squared error (MSE). For multi-task classification, we optimize the average loss across tasks. When labels are missing for some tasks in multi-task classification, we compute the loss only on observed labels.

## B.4 Optimization and Training Details

We use a common downstream training setup across all models, following [Ross et al., 2022]. Optimization is performed with AdamW using a learning rate of $3 \times 10^{-5}$ and a polynomial decay learning-rate schedule (no warmup). We train for up to 500 epochs and apply early stopping if the validation loss does not improve for 10 consecutive epochs. AdamW hyperparameters are $\beta_1 = 0.9$ and $\beta_2 = 0.98$.

Batch size is chosen based on the dataset size to balance performance and computational efficiency: 32 if the number of samples is $\leq 1000$, 256 for 1001−5000 samples, and 512 for $> 5000$ samples. All downstream experiments are conducted on a single NVIDIA H100 GPU.

# C Alternative Metrics to Pre-training Loss: Definitions and Computation

## C.1 Common Protocol

To better characterize downstream trends that cannot be captured by pre-training loss alone, we evaluate two metrics beyond pre-training loss for predicting downstream performance: the trace of the Hessian ($\mathrm{Tr}(\boldsymbol{H})$) and the Principal Gradient-based Measurement (PGM) distance.

To compute these metrics, we keep model size and training data size fixed and evaluate them on a sequence of 16 checkpoints saved at 0.25-epoch intervals during pre-training. Let the parameters of the $i$-th checkpoint be $\boldsymbol{\theta}^{(i)}$, and denote the metric value by $m(\boldsymbol{\theta}^{(i)})$.

| Category | Benchmark | Dataset | Task type | Metrics | #samples | #tasks | Data split | Vocab size |
|---|---|---|---|---|---|---|---|---|
| Physical chemistry | ESOL | ESOL | regression | RMSE | 1,127 | 1 | Random | 29 |
| | FreeSolv | FreeSolv | regression | RMSE | 641 | 1 | Random | 26 |
| | Lipophilicity | Lipophilicity | regression | RMSE | 4200 | 1 | Random | 33 |
| Biophysics | HIV | HIV | classification | ROC AUC | 41,126 | 1 | Scaffold | 188 |
| | BACE | BACE | classification | ROC AUC | 1,512 | 1 | Scaffold | 30 |
| | $\text{BACE}_{\text{IC50}}$ | | regression | RMSE | 1,512 | 1 | Scaffold | 30 |
| Physiology | BBBP | BBBP | classification | ROC AUC | 2,038 | 1 | Random | 51 |
| | Tox21 | Tox21 | classification | ROC AUC | 7,830 | 12 | Random | 132 |
| | SR-p53 | | classification | ROC AUC | 6,773 | 1 | Random | 127 |
| | SIDER | SIDER | classification | ROC AUC | 1,426 | 27 | Random | 103 |
| | ClinTox | ClinTox | classification | ROC AUC | 1,476 | 2 | Random | 69 |
| | Clearance | Clearance | regression | RMSE | 837 | 1 | Random | 30 |
| Quantum mechanics | E1-CC2 | QM8 | regression | MAE | 21,747 | 1 | Random | 31 |
| | E2-CC2 | | | | 21,747 | 1 | Random | 31 |
| | f1-CC2 | | | | 21,747 | 1 | Random | 31 |
| | f2-CC2 | | | | 21,747 | 1 | Random | 31 |
| | E1-PBE0 | | | | 21,747 | 1 | Random | 31 |
| | E2-PBE0 | | | | 21,747 | 1 | Random | 31 |
| | f1-PBE0 | | | | 21,747 | 1 | Random | 31 |
| | f2-PBE0 | | | | 21,747 | 1 | Random | 31 |
| | E1-CAM | | | | 21,747 | 1 | Random | 31 |
| | E2-CAM | | | | 21,747 | 1 | Random | 31 |
| | f1-CAM | | | | 21,747 | 1 | Random | 31 |
| | f2-CAM | | | | 21,747 | 1 | Random | 31 |
| | $\alpha$ | QM9 | regression | MAE | 133,885 | 1 | Random | 30 |
| | $C_v$ | | | | 133,885 | 1 | Random | 30 |
| | G | | | | 133,885 | 1 | Random | 30 |
| | Gap | | | | 133,885 | 1 | Random | 30 |
| | H | | | | 133,885 | 1 | Random | 30 |
| | $\varepsilon_{homo}$ | | | | 133,885 | 1 | Random | 30 |
| | $\varepsilon_{lumo}$ | | | | 133,885 | 1 | Random | 30 |
| | $\mu$ | | | | 133,885 | 1 | Random | 30 |
| | $\langle R^2 \rangle$ | | | | 133,885 | 1 | Random | 30 |
| | $U_0$ | | | | 133,885 | 1 | Random | 30 |
| | $U$ | | | | 133,885 | 1 | Random | 30 |
| | ZPVE | | | | 133,885 | 1 | Random | 30 |

Table 3: Downstream benchmarks used in this study. For each benchmark, we report the task type, evaluation metric, number of samples and tasks, split type, and vocabulary size.

Hereafter, we interpret each metric so that smaller values correspond to better downstream performance. To match this direction, we also express downstream performance $P$ in a lower-is-better form: for classification tasks, we use $P = 1 -$ ROC-AUC; for regression tasks, we use MAE or RMSE as-is. For both $P^{\text{FT}}$ and $P^{\text{LP}}$, we evaluate the strength of the monotonic relationship between metric values and downstream performance using Spearman's rank correlation coefficient, denoted by $\rho$.

## C.2 Trace of the Hessian

**Definition**. For each checkpoint $\boldsymbol{\theta}$, we consider the Hessian of the pre-training loss $L_{\text{pre}}(\boldsymbol{\theta})$ and use its trace $\text{Tr}(\boldsymbol{H})$ as a quantity that captures the local curvature of the loss landscape. In general, a smaller $\text{Tr}(\boldsymbol{H})$ indicates a flatter landscape, which has been reported to be related to downstream performance in prior work. In this study, we compute $\text{Tr}(\boldsymbol{H})$ only with respect to the encoder parameters.

**Estimation**. Since computing $\text{Tr}(\boldsymbol{H})$ exactly is infeasible for large models, we use Hutchinson's trace estimator. For a random vector $\boldsymbol{v}$ with mean $\boldsymbol{0}$ and covariance $\boldsymbol{I}$, we have:

$$\mathbb{E}[\boldsymbol{v}^{\text{T}}\boldsymbol{H}\boldsymbol{v}] = \text{Tr}(\boldsymbol{H}).$$

Using $M$ samples $\{\boldsymbol{v}_k\}_{k=1}^{M}$, we estimate $\text{Tr}(\boldsymbol{H})$ by

$$\widehat{\mathrm{Tr}}(\boldsymbol{H}) = \frac{1}{M}\sum_{k=1}^{M} \boldsymbol{v}_k^{\mathrm{T}} \boldsymbol{H} \boldsymbol{v}_k.$$

In this study, we sample $\boldsymbol{v}_k \sim \mathcal{N}(\boldsymbol{0}, \boldsymbol{I})$.

**HVP computation**. Each term $\boldsymbol{v}_k^{\mathrm{T}} \boldsymbol{H} \boldsymbol{v}_k$ is computed using a Hessian-vector product (HVP). Let the mini-batch loss be $L(\boldsymbol{\theta})$ and its gradient be $\boldsymbol{g} = \nabla_{\boldsymbol{\theta}} L(\boldsymbol{\theta})$. Then,

$$\boldsymbol{H}\boldsymbol{v}_k = \nabla_{\boldsymbol{\theta}}(\boldsymbol{g}^{\mathrm{T}} \boldsymbol{v}_k).$$

Which can be computed via automatic differentiation. Finally, we obtain $\boldsymbol{v}_k^{\mathrm{T}}(\boldsymbol{H}\boldsymbol{v}_k)$ to compute $\boldsymbol{v}_k^{\mathrm{T}} \boldsymbol{H} \boldsymbol{v}_k$.

**Implementation details**. For each checkpoint $\boldsymbol{\theta}$, we randomly sample 50,000 input sequences from the pre-training data and split them into mini-batches of size 128. For each mini-batch, we sample one vector $\boldsymbol{v}$ and compute $\boldsymbol{v}^{\mathrm{T}}\boldsymbol{H}\boldsymbol{v}$. Therefore, $M$ equals the number of mini-batches. The sampled input sequences are resampled for each checkpoint.

### C.3 Principal Gradient-based Measurement (PGM) Distance

PGM distance is a proxy metric that approximates transferability without any additional optimization, based on the discrepancy between the gradient directions induced by a source task $T_s$ and a target task $T_t$ [Yao et al., 2024]. In this study, we set $T_s$ to MLM and $T_t$ to each downstream task.

**Principal gradient**. For a checkpoint $\boldsymbol{\theta}$, we define the principal gradient of task $T$ as the mean gradient over $K$ independently sampled mini-batches:

$$\bar{g}_T(\boldsymbol{\theta}) = \frac{1}{K}\sum_{k=1}^{K} \nabla_{\boldsymbol{\theta}} L_T(\boldsymbol{\theta}; B_k).$$

where $B_k$ denotes a mini-batch sampled from the dataset of task $T$ and $L_T$ is the corresponding task loss. To reduce the influence of outlier gradients that may arise from any single mini-batch, we average gradients across multiple forward/backward passes. Considering the trade-off with computational cost, we set $K = 10$ in this study.

**PGM distance**. We define PGM distance as the normalized distance between the principal gradients of the source and target tasks:

$$\mathrm{PGMdist}_{T_s \to T_t}(\boldsymbol{\theta}) = \frac{\left\| \bar{g}_{T_t}(\boldsymbol{\theta}) - \bar{g}_{T_s}(\boldsymbol{\theta}) \right\|_2}{\left\| \bar{g}_{T_s}(\boldsymbol{\theta}) \right\|_2 \left\| \bar{g}_{T_t}(\boldsymbol{\theta}) \right\|_2}.$$

We interpret smaller values as indicating that the two tasks induce more aligned gradient directions, implying easier transfer.

## D Scaling-law Fitting Details

### D.1 Purpose

This section summarizes the fitting procedure used to quantify the pre-training loss scaling behavior shown in the main text with a simple parametric model. The purpose is not to claim universality of the estimated parameters. Rather, we aim to confirm that, within the ranges of model size and data size considered in our experiments, increasing training resources is consistently associated with improved pre-training loss.

### D.2 Parametric Model of the Final Loss

Following the parametric fitting approach of [Hoffmann et al., 2022], we model the validation loss at the final checkpoint of each pre-training condition as a function of model size $N$ and data size $D$.

$$\hat{L}(N, D) \triangleq E + \frac{A}{N^{\alpha}} + \frac{B}{D^{\beta}}$$

Here, $E$ represents an irreducible loss floor due to the data-generating process; the $N$-dependent term captures approximation error from finite model capacity; and the $D$-dependent term captures optimization error arising from finite training data and finite training steps. We define $N$ as the number of non-embedding parameters and $D$ as the number of training tokens used in the corresponding experiment.

### D.3 Fitting Objective

We estimate the parameters $A, B, E, \alpha, \beta$ by minimizing the Huber loss on the discrepancy between the log observed loss and the log predicted loss. Specifically, letting $L_i$ denote the observed final validation loss for data point $i$, we solve:

$$\min_{A,B,E,\alpha,\beta} \sum_i \mathrm{Huber}_{\delta}\left(\log \hat{L}(N_i, D_i) - \log L_i\right)$$

Because the Huber threshold $\delta$ affects robustness to outliers, we sweep $\delta$ from $10^{-2}$ to $10^{-6}$, compare $R^2$, and adopt the best setting. In our study, $\delta = 10^{-3}$ performed best.

### D.4 Optimization

We use L-BFGS for optimization. For numerical stability, following [Hoffmann et al., 2022], we optimize $A, B, E$ in log-space by introducing variables $a, b, e$ such that $A = \exp(a)$, $B = \exp(b)$, and $E = \exp(e)$. We also stabilize the computation of $\hat{L}$ using log-sum-exp trick, which helps avoid underflow/overflow at extreme scales.

### D.5 Data Points Used for Fitting

We fit the scaling model using the final losses obtained from multiple $(N, D)$ pairs. However, as shown in Figure 8, under the small-data conditions (0.7 and 1.3 billion tokens), training ended before the loss had converged, and including these points together with converged points made the fit unstable. Following [Frey et al., 2023], we therefore exclude these conditions and fit the model using 21 points. The exclusion is based on observations such as the loss continuing to decrease noticeably near the end of training without clear saturation.

### D.6 Fitting Results and Interpretation

With the above setup, we obtain $\alpha = 0.686$, $\beta = 1.702$, $A = 1.545 \times 10^{3}$, $B = 4.250 \times 10^{14}$, $E = 2.880 \times 10^{-2}$, achieving a high goodness of fit with $R^2 = 0.989$. Although the estimated exponents $\alpha$ and $\beta$ exceed an upper bound of 0.5 on the exponent discussed in some prior work [Robbins and Monro, 1951; Siegel and Xu, 2020], we do not claim universality of these coefficients. The key point is that, within the ranges of model size and data size considered in this study, increasing $N$ and $D$ is quantitatively associated with a consistent reduction in loss under this simple model.

| Methods | BBBP | Tox21 | ClinTox | HIV | BACE | SIDER | SR-p53 |
|---|---|---|---|---|---|---|---|
| Uni-Mol [Zhou et al., 2023] | 71.5 | 78.9 | 84.1 | 78.6 | 83.2 | 57.7 | - |
| MoleBlend [Yu et al., 2024] | 73.0 | 77.8 | 87.6 | 79.0 | 83.7 | 64.9 | - |
| Mol-AE [Yang et al., 2024] | 72.0 | 80.0 | 87.8 | 80.6 | 84.1 | 67.0 | - |
| UniCorn [Feng et al., 2024] | 74.2 | 79.3 | 92.1 | 79.8 | 85.8 | 64.0 | - |
| SimSGT [Liu et al., 2023] | 72.2 | 76.8 | 85.7 | 78.0 | 84.3 | 61.7 | - |
| SELFormer [Yüksel et al., 2023] | 90.2 | 65.3 | - | 68.1 | 83.2 | 74.5 | - |
| MolTRES [Park et al., 2024] | 96.1 | 85.3 | 96.7 | 84.2 | 91.7 | 69.8 | - |
| ChemBERTa [Chithrananda et al., 2020] | 90.3 | 79.1 | 91.7 | 71.9 | 85.7 | 64.3 | 85.2 |
| MolFormer-XL [Ross et al., 2022] | 93.7 | 84.7 | 94.8 | 82.2 | 88.2 | 69.0 | 83.5 |
| **Ours (0.25M)** | 72.7 | 58.1 | 48.8 | 60.0 | 62.3 | 51.8 | 66.8 |
| **Ours (1.20M)** | 89.6 | 66.3 | 57.0 | 76.7 | 85.5 | 57.0 | 77.2 |
| **Ours (2.86M)** | 93.5 | 80.4 | 76.9 | 78.0 | 86.9 | 58.6 | 80.8 |
| **Ours (5.61M)** | 92.3 | 80.8 | 80.2 | 79.2 | 86.1 | 59.3 | 84.9 |
| **Ours (27.23M)** | 93.2 | 81.9 | 92.0 | 81.3 | 87.5 | 65.9 | 85.7 |
| **Ours (88.46M)** | 93.1 | 82.9 | 90.2 | 81.8 | 87.3 | 65.5 | 88.0 |
| **Ours (206.60M)** | 93.5 | 83.7 | 94.0 | 84.0 | 87.2 | 65.8 | 86.5 |

Table 4: Classification benchmark results (ROC-AUC, ↑). “-” denotes results not reported in prior work.

# E Compute Metric

## E.1 Overview

In this study, we report the compute used for pre-training in PF-days (petaflop-days). PF-days expresses total compute in terms of FLOPs (floating-point operations) and normalizes by the compute of running 1 PFLOP/s for one day, i.e., $10^{15} \times 86400$. This provides a single scale that is comparable across experimental conditions with different model sizes and training durations. Following the definition in [Kaplan et al., 2020], we consistently estimate pre-training compute using theoretical FLOPs-based formulas rather than measured GPU wall-clock time.

## E.2 Calculating FLOPs

Following the approximation in [Kaplan et al., 2020], let $N$ denote the number of non-embedding parameters. The compute required to process one token during training is approximated as

$$\mathrm{FLOPs}_{\mathrm{token}} \approx 6N.$$

Here, the coefficient 6 approximates the dominant training compute including both forward and backward passes; the actual compute can vary depending on implementation details. To avoid variability due to such implementation differences, we report compute for all experimental conditions using this unified theoretical formula. Let $D_{\mathrm{train}}$ be the total number of tokens processed during training. Then the total FLOPs is given by

$$\mathrm{FLOPs}_{\mathrm{total}} = \mathrm{FLOPs}_{\mathrm{token}} \times D_{\mathrm{train}} \approx 6ND_{\mathrm{train}}.$$

Equivalently, if $B_{\mathrm{token}}$ denotes the number of tokens processed per training step and $S$ denotes the number of training steps, then $D_{\mathrm{train}} = B_{\mathrm{token}}S$, and we obtain

$$\mathrm{FLOPs}_{\mathrm{total}} \approx 6NB_{\mathrm{token}}S.$$

## E.3 Calculating PF-days

PF-days is computed from the total FLOPs as

$$\text{PF-days} = \frac{\mathrm{FLOPs}_{\mathrm{total}}}{10^{15} \times 86400}.$$

In this study, we calculate $D_{\mathrm{train}}$ from $B_{\mathrm{token}}$ and the cumulative number of steps $S$ for each training run, and report the pre-training compute cost of each model using the above formula.

# F Comparison with Prior Models on Downstream Benchmarks

In this section, we provide a reference comparison of downstream benchmark performance for existing molecular property prediction (MPP) models against the pre-trained models studied in this work (Tables 4–9). Although our primary goal is not to develop the highest-performing MPP model, we present these comparisons to provide context for how our pre-trained models perform on downstream tasks.

For our models, we use checkpoints after 4 epochs of training with the largest pre-training data setting, and report downstream performance for each of the seven model sizes. Downstream evaluation follows the protocol described in Appendix B, and we report fine-tuning.

For prior models, we primarily cite the values reported in the original papers. If a value is not reported, we denote it with “-”. In addition, for MolFormer-XL and ChemBERTa, we re-train and re-evaluate these baselines on a subset of tasks using the same protocol as Appendix B, and use resulting scores to fill missing entries in the tables.

| Methods | ESOL | FreeSolv | Lipophilicity | Clearance | $BACE_{IC50}$ |
|---|---|---|---|---|---|
| Uni-Mol [Zhou et al., 2023] | 0.844 | 1.879 | 0.610 | - | - |
| MoleBlend [Yu et al., 2024] | 0.831 | 1.910 | 0.638 | - | - |
| Mol-AE [Yang et al., 2024] | 0.830 | 1.448 | 0.607 | - | - |
| UniCorn [Feng et al., 2024] | 0.817 | 1.555 | 0.591 | - | - |
| SimSGT [Liu et al., 2023] | 0.917 | - | 0.695 | - | - |
| SELFormer [Yüksel et al., 2023] | 0.682 | 2.797 | 0.735 | - | - |
| MolTRES [Park et al., 2024] | 0.274 | 0.229 | 0.504 | - | - |
| ChemBERTa [Chithrananda et al., 2020] | 0.695 | 0.397 | 0.629 | 51.311 | 0.826 |
| MolFormer-XL [Ross et al., 2022] | 0.279 | 0.231 | 0.529 | 45.400 | 0.771 |
| **Ours (0.25M)** | 1.608 | 0.881 | 0.994 | 52.223 | 1.435 |
| **Ours (1.20M)** | 0.727 | 0.372 | 0.637 | 46.445 | 0.879 |
| **Ours (2.86M)** | 0.625 | 0.339 | 0.590 | 45.956 | 0.875 |
| **Ours (5.61M)** | 0.620 | 0.321 | 0.589 | 44.872 | 0.850 |
| **Ours (27.23M)** | 0.600 | 0.278 | 0.541 | 44.406 | 0.833 |
| **Ours (88.46M)** | 0.565 | 0.257 | 0.512 | 43.044 | 0.811 |
| **Ours (206.60M)** | 0.538 | 0.255 | 0.523 | 42.590 | 0.726 |

Table 5: Non-quantum regression benchmark results (RMSE, ↓). "-" denotes results not reported in prior work.

| Methods | $\alpha$ | $C_v$ | G | gap | H | $\varepsilon_{homo}$ |
|---|---|---|---|---|---|---|
| A-FP [Xiong et al., 2020] | 0.492 | 0.252 | 0.893 | 0.0053 | 0.893 | 0.0036 |
| 123-gnn [Morris et al., 2019] | 0.27 | 0.0944 | 0.0469 | 0.0048 | 0.0419 | 0.0034 |
| GC [Altae-Tran et al., 2017] | 1.37 | 0.65 | 3.41 | 0.0113 | 3.41 | 0.0072 |
| CM [Rupp et al., 2012] | 0.85 | 0.39 | 2.27 | 0.0086 | 2.27 | 0.0051 |
| DTNN [Schütt et al., 2017] | 0.95 | 0.27 | 2.43 | 0.0112 | 2.43 | 0.0038 |
| MPNN [Gilmer et al., 2017] | 0.89 | 0.42 | 2.02 | 0.0066 | 2.02 | 0.0054 |
| ChemBERTa [Chithrananda et al., 2020] | 0.6658 | 0.2644 | 2.1753 | 0.0053 | 2.0175 | 0.0041 |
| MolFormer-XL [Ross et al., 2022] | 0.3327 | 0.1447 | 0.3362 | 0.0038 | 0.2522 | 0.0029 |
| **Ours (0.25M)** | 4.9131 | 1.0165 | 6.9919 | 0.0155 | 7.1197 | 0.0098 |
| **Ours (1.20M)** | 0.7231 | 0.3291 | 1.6934 | 0.0074 | 1.6082 | 0.0048 |
| **Ours (2.86M)** | 0.5465 | 0.2254 | 0.8938 | 0.0062 | 0.8488 | 0.0042 |
| **Ours (5.61M)** | 0.4763 | 0.1428 | 0.5691 | 0.0054 | 0.6984 | 0.0037 |
| **Ours (27.23M)** | 0.3618 | 0.1749 | 0.5539 | 0.0045 | 0.3695 | 0.0033 |
| **Ours (88.46M)** | 0.2875 | 0.1637 | 0.2373 | 0.0042 | 0.2477 | 0.0030 |
| **Ours (206.60M)** | 0.2853 | 0.1212 | 0.2973 | 0.0039 | 0.2493 | 0.0027 |

Table 6: QM9 quantum chemistry regression benchmark results (MAE, ↓) (part 1/2).

| Methods | $\varepsilon_{lumo}$ | $\mu$ | $\langle R^2 \rangle$ | $U_0$ | $U$ | ZPVE |
|---|---|---|---|---|---|---|
| A-FP [Xiong et al., 2020] | 0.0042 | 0.451 | 26.839 | 0.898 | 0.893 | 0.0021 |
| 123-gnn [Morris et al., 2019] | 0.0035 | 0.476 | 22.90 | 0.0427 | 0.111 | 0.0002 |
| GC [Altae-Tran et al., 2017] | 0.0092 | 0.583 | 35.97 | 3.41 | 3.41 | 0.0030 |
| CM [Rupp et al., 2012] | 0.0065 | 0.519 | 46.00 | 2.27 | 2.27 | 0.0021 |
| DTNN [Schütt et al., 2017] | 0.0051 | 0.244 | 17.00 | 2.43 | 2.43 | 0.0017 |
| MPNN [Gilmer et al., 2017] | 0.0062 | 0.358 | 28.5 | 2.05 | 2 | 0.0022 |
| ChemBERTa [Chithrananda et al., 2020] | 0.0048 | 0.4916 | 30.8104 | 1.9501 | 1.8529 | 0.0017 |
| MolFormer-XL [Ross et al., 2022] | 0.0027 | 0.3616 | 17.0620 | 0.3211 | 0.2522 | 0.0003 |
| **Ours (0.25M)** | 0.0160 | 0.8013 | 151.1001 | 6.9221 | 6.8818 | 0.0081 |
| **Ours (1.20M)** | 0.0057 | 0.5428 | 32.8202 | 1.6507 | 1.6184 | 0.0022 |
| **Ours (2.86M)** | 0.0044 | 0.4978 | 27.3829 | 0.7772 | 0.7429 | 0.0012 |
| **Ours (5.61M)** | 0.0040 | 0.4677 | 25.5322 | 0.6032 | 0.6722 | 0.0013 |
| **Ours (27.23M)** | 0.0035 | 0.4351 | 22.0947 | 0.5049 | 0.4518 | 0.0009 |
| **Ours (88.46M)** | 0.0031 | 0.4051 | 19.1873 | 0.3432 | 0.2251 | 0.0007 |
| **Ours (206.60M)** | 0.0029 | 0.3882 | 19.0943 | 0.2628 | 0.2354 | 0.0004 |

Table 7: QM9 quantum chemistry regression benchmark results (MAE, ↓ ) (part 2/2).

| Methods | E1-CC2 | E2-CC2 | f1-CC2 | f2-CC2 | E1-PBE0 | E2-PBE0 |
|---|---|---|---|---|---|---|
| ChemBERTa [Chithrananda et al., 2020] | 0.0307 | 0.0213 | 0.0241 | 0.0433 | 0.0326 | 0.0252 |
| MolFormer-XL [Ross et al., 2022] | 0.0301 | 0.0205 | 0.0239 | 0.0443 | 0.0321 | 0.0245 |
| **Ours (0.25M)** | 0.0357 | 0.0244 | 0.0295 | 0.0488 | 0.0387 | 0.0304 |
| **Ours (1.20M)** | 0.0309 | 0.0205 | 0.0250 | 0.0458 | 0.0320 | 0.0250 |
| **Ours (2.86M)** | 0.0296 | 0.0203 | 0.0241 | 0.0448 | 0.0311 | 0.0242 |
| **Ours (5.61M)** | 0.0291 | 0.0202 | 0.0243 | 0.0447 | 0.0307 | 0.0236 |
| **Ours (27.23M)** | 0.0300 | 0.0203 | 0.0237 | 0.0448 | 0.0313 | 0.0242 |
| **Ours (88.46M)** | 0.0300 | 0.0204 | 0.0235 | 0.0446 | 0.0315 | 0.0243 |
| **Ours (206.60M)** | 0.0302 | 0.0205 | 0.0228 | 0.0436 | 0.0319 | 0.0246 |

Table 8: QM8 quantum chemistry regression benchmark results (MAE, ↓ ) (part 1/2).

| Methods | f1-PBE0 | f2-PBE0 | E1-CAM | E2-CAM | f1-CAM | f2-CAM |
|---|---|---|---|---|---|---|
| ChemBERTa [Chithrananda et al., 2020] | 0.0229 | 0.0337 | 0.0309 | 0.0218 | 0.0273 | 0.0383 |
| MolFormer-XL [Ross et al., 2022] | 0.0220 | 0.0328 | 0.0296 | 0.0218 | 0.0254 | 0.0383 |
| **Ours (0.25M)** | 0.0305 | 0.0379 | 0.0362 | 0.0261 | 0.0343 | 0.0442 |
| **Ours (1.20M)** | 0.0239 | 0.0363 | 0.0307 | 0.0214 | 0.0276 | 0.0412 |
| **Ours (2.86M)** | 0.0230 | 0.0356 | 0.0291 | 0.0210 | 0.0263 | 0.0394 |
| **Ours (5.61M)** | 0.0229 | 0.0352 | 0.0293 | 0.0206 | 0.0267 | 0.0390 |
| **Ours (27.23M)** | 0.0221 | 0.0353 | 0.0299 | 0.0211 | 0.0253 | 0.0398 |
| **Ours (88.46M)** | 0.0215 | 0.0353 | 0.0302 | 0.0211 | 0.0252 | 0.0399 |
| **Ours (206.60M)** | 0.0218 | 0.0346 | 0.0297 | 0.0213 | 0.0243 | 0.0385 |

Table 9: QM8 quantum chemistry regression benchmark results (MAE, ↓ ) (part 2/2).

# G Task-wise Scaling Curves for 36 Benchmarks

In this appendix, we present task-wise changes in downstream performance as pre-training resources are scaled across 36 benchmarks. While Figure 3 in the main paper summarizes overall trends by averaging over three task groups (classification, non-quantum regression, and quantum-chemistry regression), here we report results for individual tasks. The benchmarks include 7 classification tasks, 5 non-quantum regression tasks, and 24 quantum-chemistry regression tasks. We use the same quantities as in Figure 3. This per-task breakdown helps reveal task-to-task differences that may be obscured by group averages.

For each benchmark, downstream performance is reported as the mean over three random seeds, and error bars indicate the corresponding standard deviation across seeds. We report downstream performance under two transfer settings, namely fine-tuning performance $P^{\mathrm{FT}}$ and linear probe performance $P^{\mathrm{LP}}$. Evaluation metrics follow the task type. We use ROC-AUC for classification, MAE for quantum-chemistry, and RMSE for non-quantum regression.

In each figure, the left y-axis shows downstream performance ($P^{\mathrm{FT}}$ and $P^{\mathrm{LP}}$), and the right y-axis shows losses ($L_{\mathrm{pre}}$ and $L_{\mathrm{down}}$). The x-axis depends on the scaling factor: the number of parameters for model-size scaling, the number of training tokens for data-size scaling, and pre-training compute measured in PF-days for compute scaling.

## G.1 Model Scaling

In the model-size scaling experiment, we fix the training data size at 42.4B tokens and train for 4 epochs. We compare seven models that differ only in model size. Figure 9 shows results for classification and non-quantum regression tasks, while Figures 10 and 11 show results for tasks from QM8 and QM9, respectively. Each subplot corresponds to one benchmark and uses the number of parameters as the x-axis. For reference, we also include fine-tuning performance of existing models, MolFormer-XL [Ross et al., 2022] and ChemBERTa [Chithrananda et al., 2020].

## G.2 Data Scaling

In the data-size scaling experiment, we fix the model size to 4.87M parameters and train for 4 epochs. We compare five training settings that differ only in training data size. Figure 12 shows results for classification and non-quantum regression tasks, while Figures 13 and 14 show results for individual tasks in QM8 and QM9, respectively. In this setting, the x-axis in each subplot is the number of training tokens. Other axes and plotted quantities are the same as in Figure 9.

## G.3 Compute Scaling

In the compute scaling experiment, we fix the model size to 4.87M parameters and the training data size to 42.4B tokens. We compare checkpoints saved every 0.25 epoch during the 4-epoch pre-training run, together with 0-step checkpoint (random initialization). Figure 15 shows results for classification and non-quantum regression tasks, while Figures 16 and 17 show results for tasks from QM8 and QM9, respectively. In this setting, the x-axis is pre-training compute measured in PF-days. Other axes and plotted quantities are the same as in Figure 9.

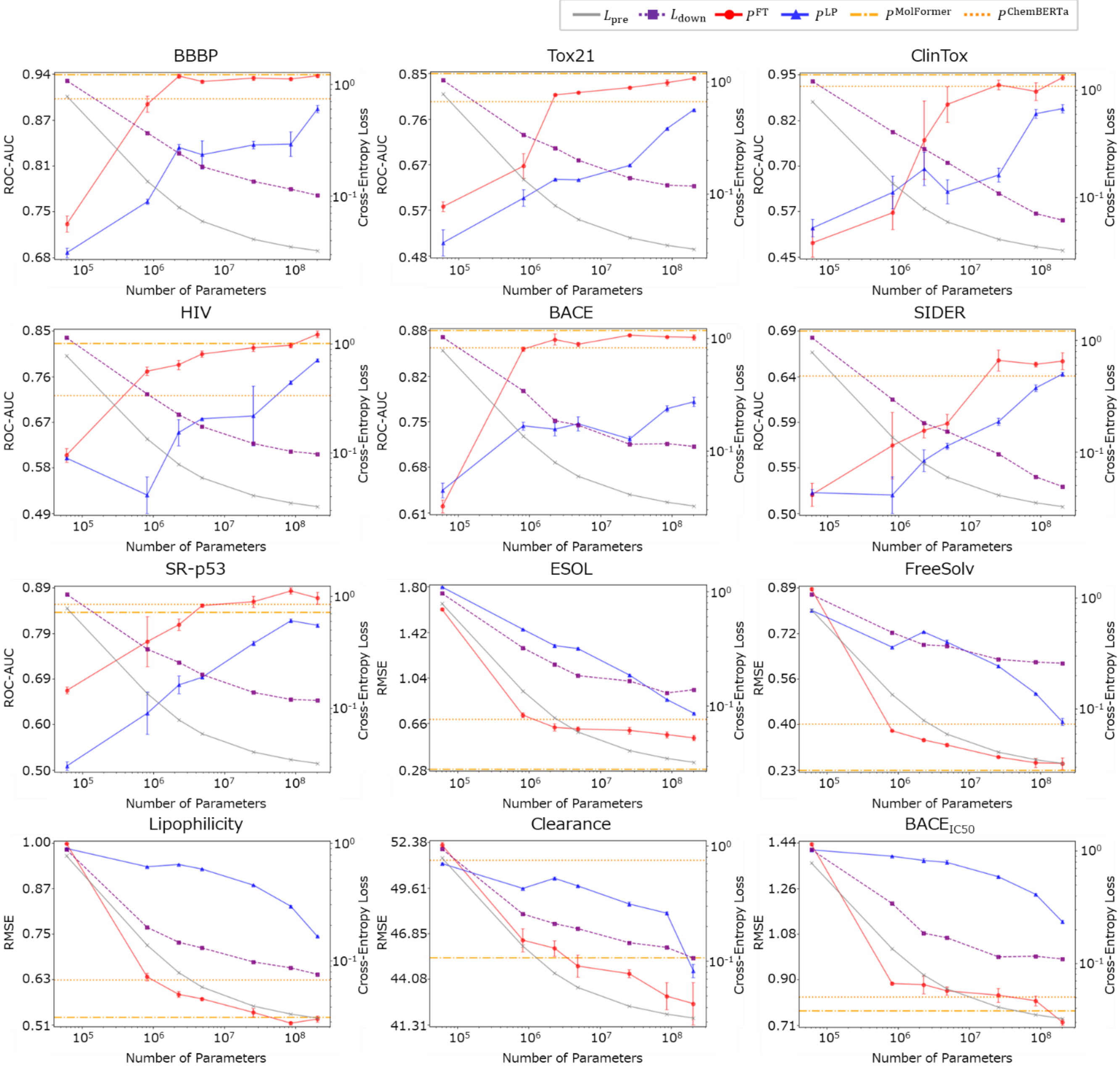

Figure 9: Task-wise curves under model-size scaling. Each subplot corresponds to a benchmark. The x-axis is the number of parameters. The left y-axis shows downstream performance ($P^{\text{FT}}$ and $P^{\text{LP}}$), and the right y-axis shows losses ($L_{\text{pre}}$ and $L_{\text{down}}$). For reference, fine-tuning performance of MolFormer-XL and ChemBERTa is shown.

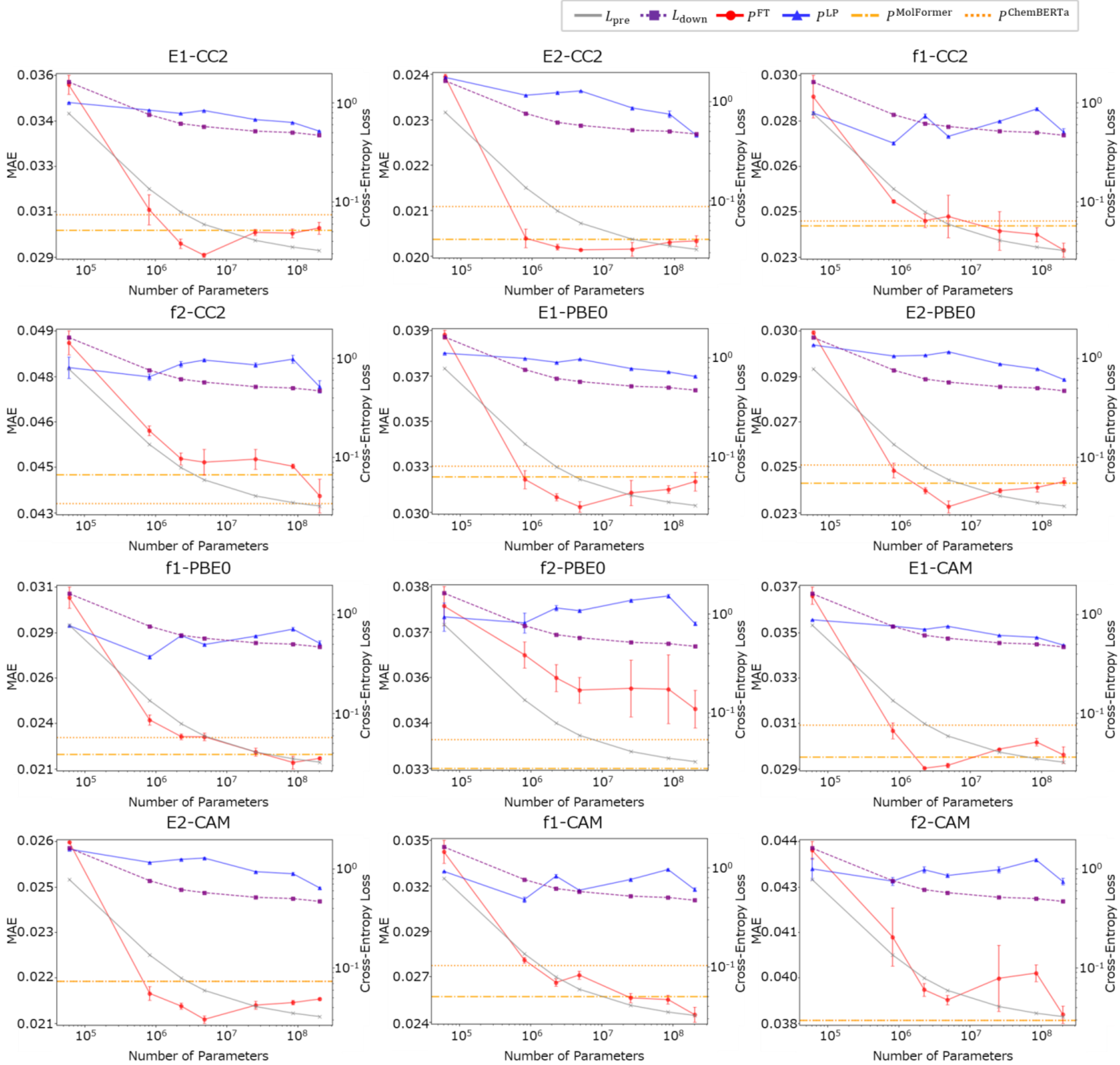

Figure 10: Task-wise curves under model-size scaling for QM8 tasks. Axes and plotted quantities are the same as Figure 9.

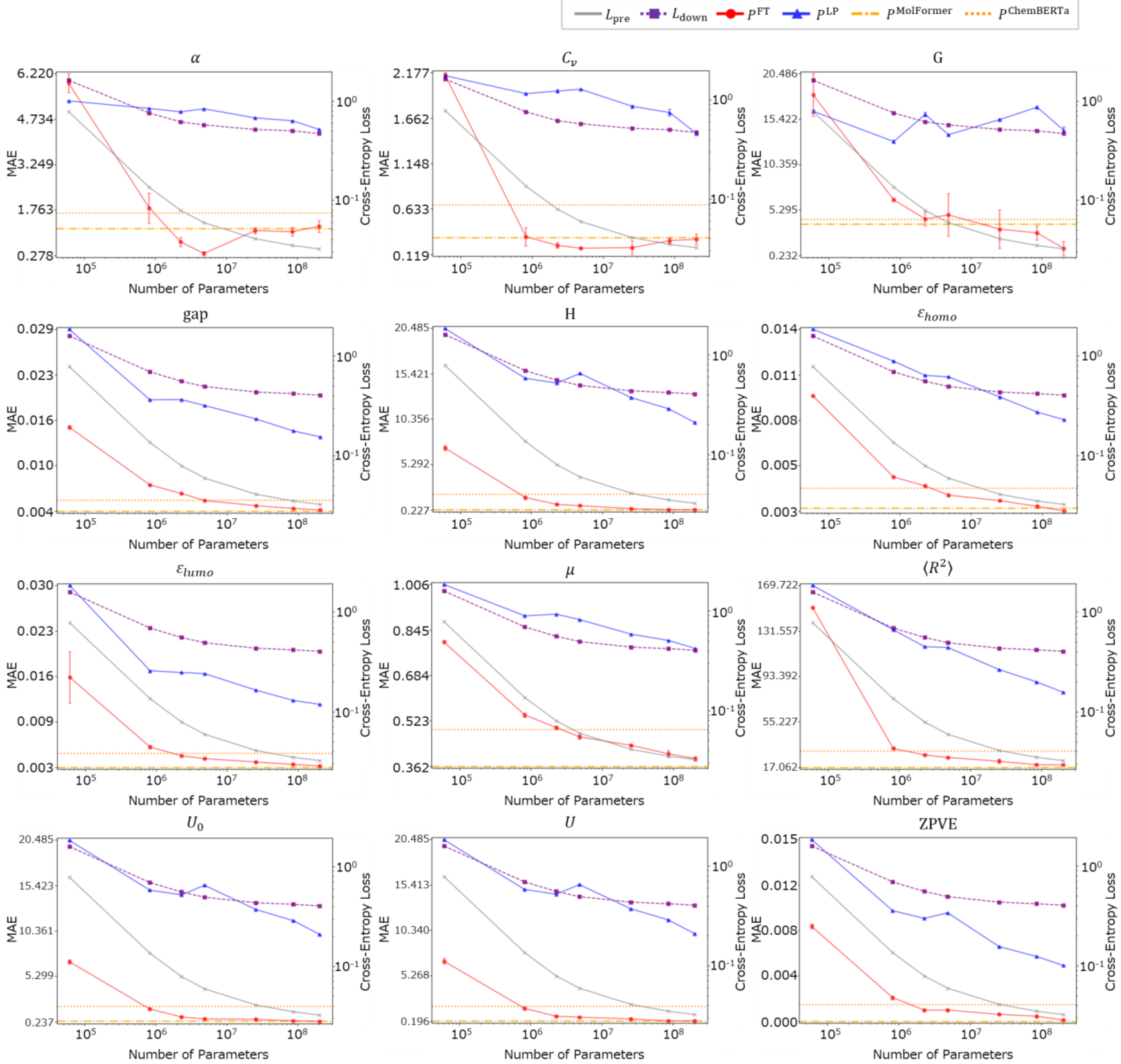

Figure 11: Task-wise curves under model-size scaling for QM9 tasks. Axes and plotted quantities are the same as Figure 9.

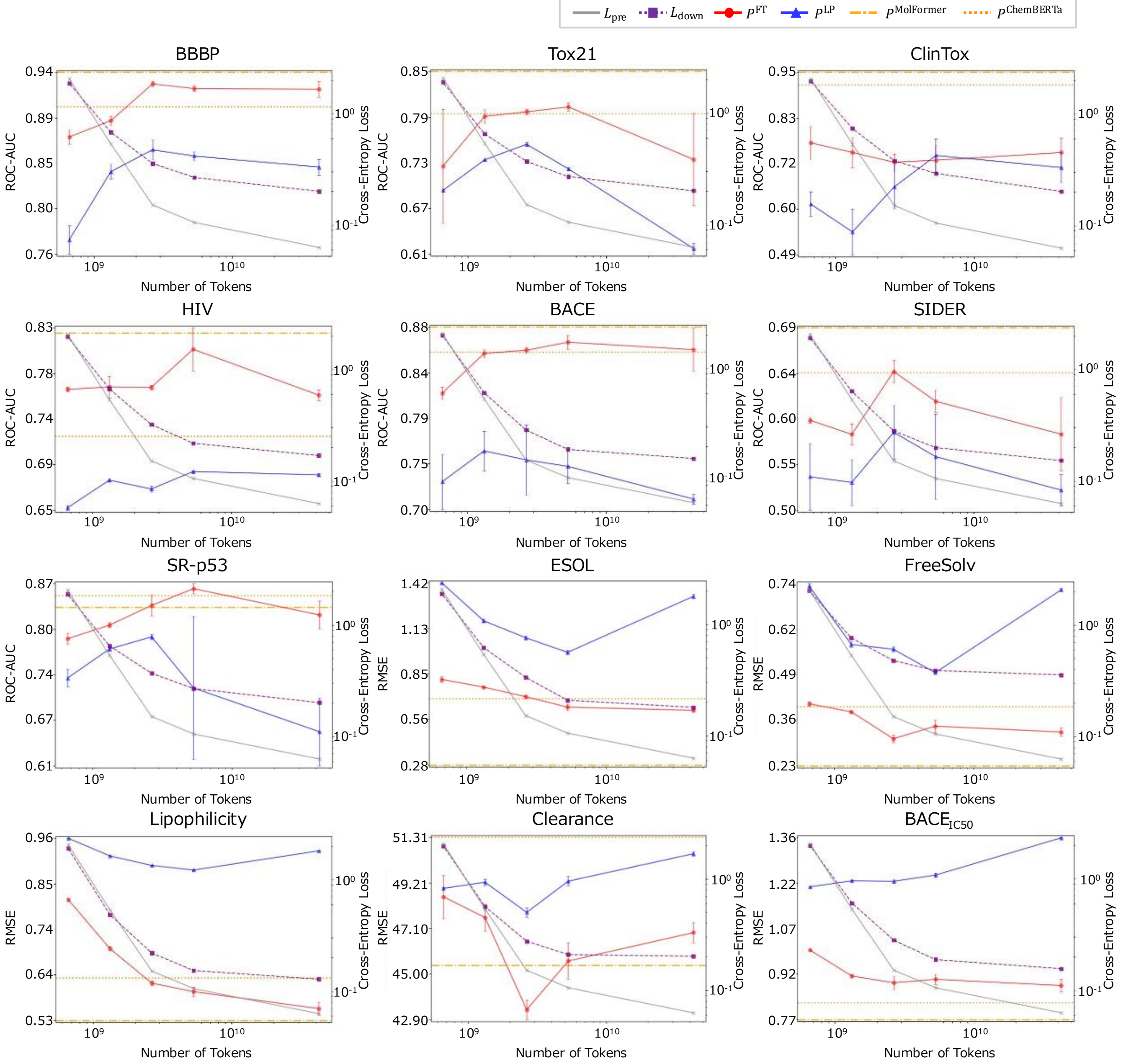

Figure 12: Task-wise curves under data-size scaling. The x-axis is the number of training tokens. Axes and plotted quantities are the same as Figure 9.

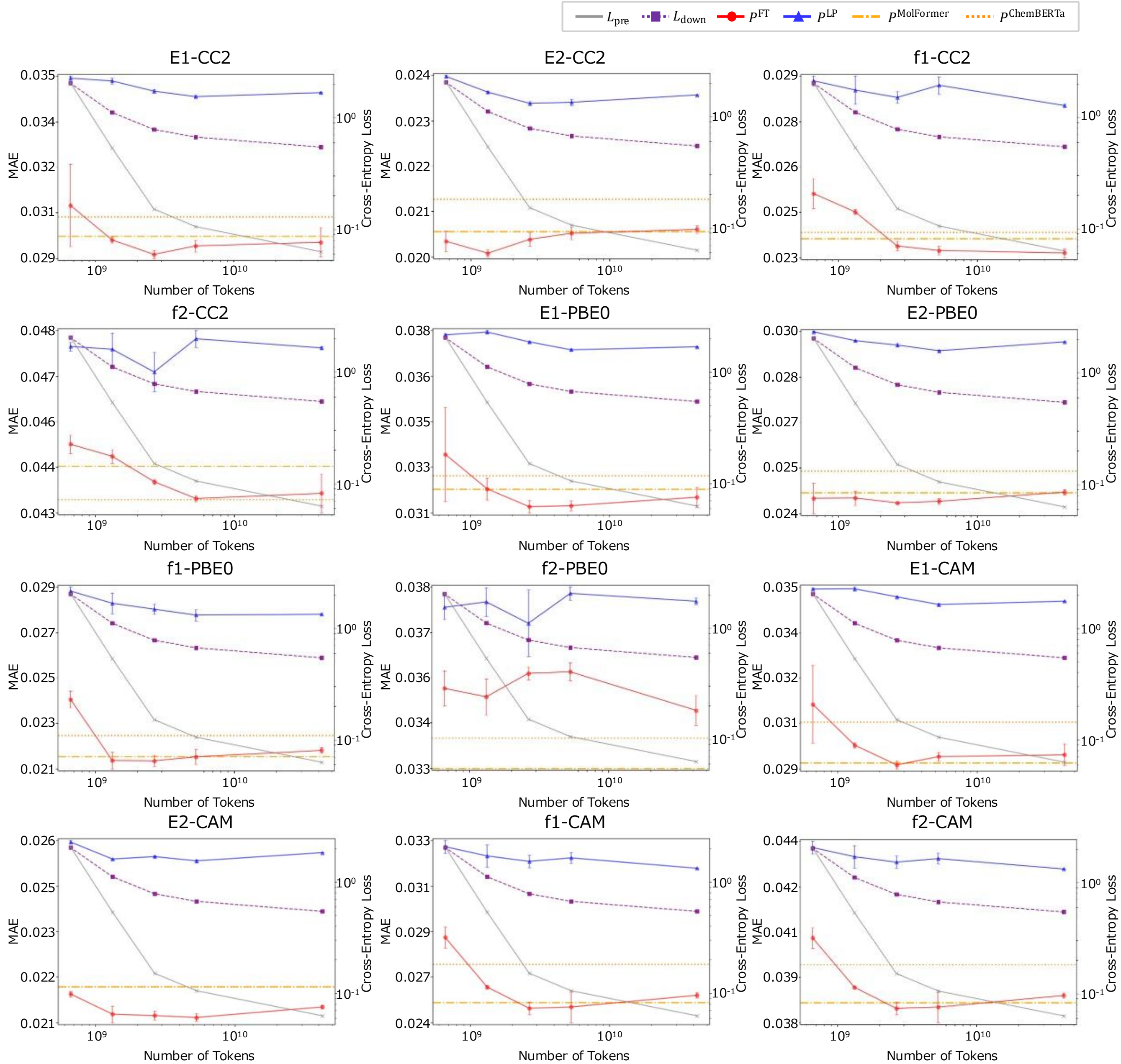

Figure 13: Task-wise curves under data-size scaling for QM8 tasks. Axes and plotted quantities are the same as Figure 12.

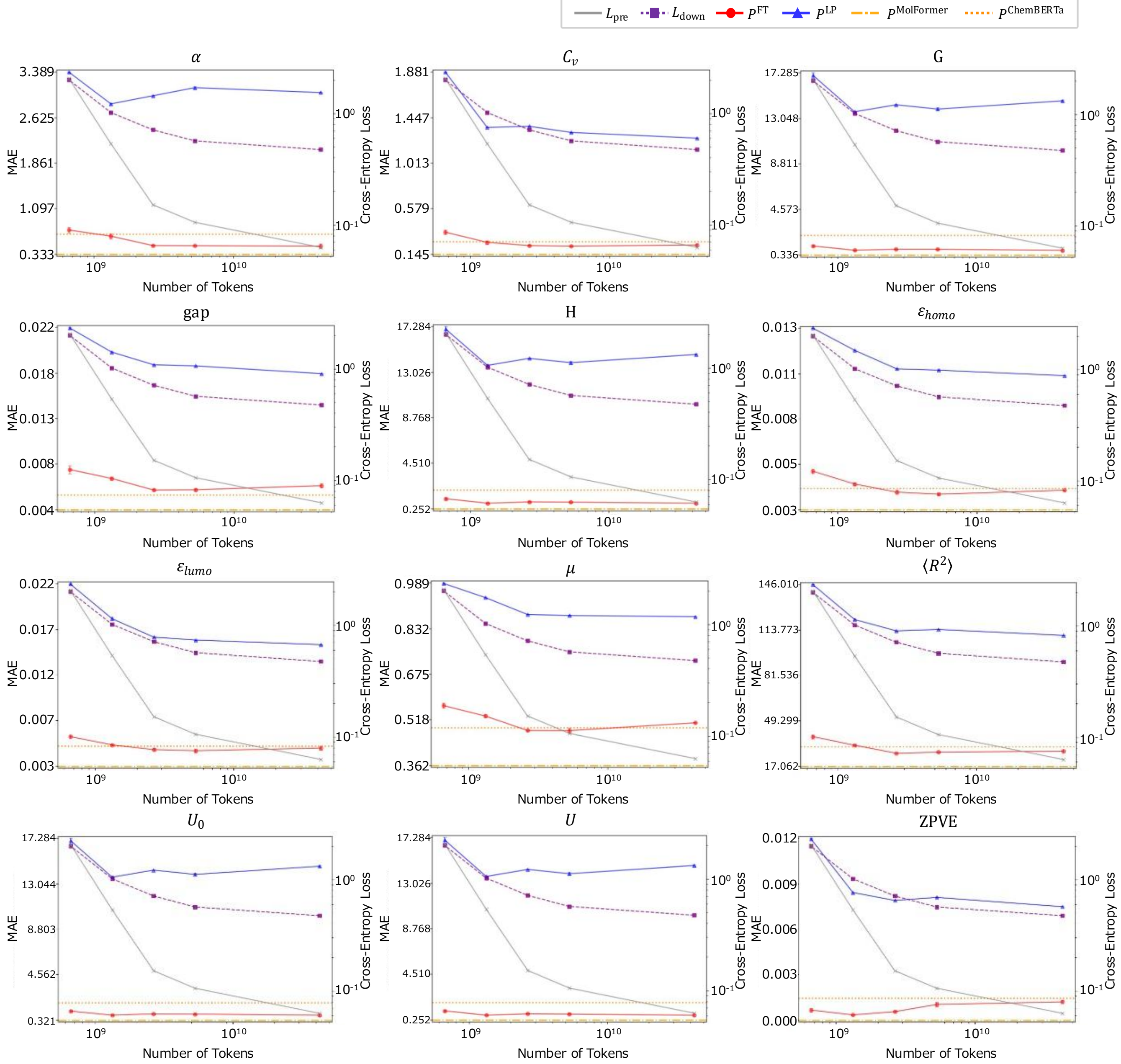

Figure 14: Task-wise curves under data-size scaling for QM9 tasks. Axes and plotted quantities are the same as Figure 12.

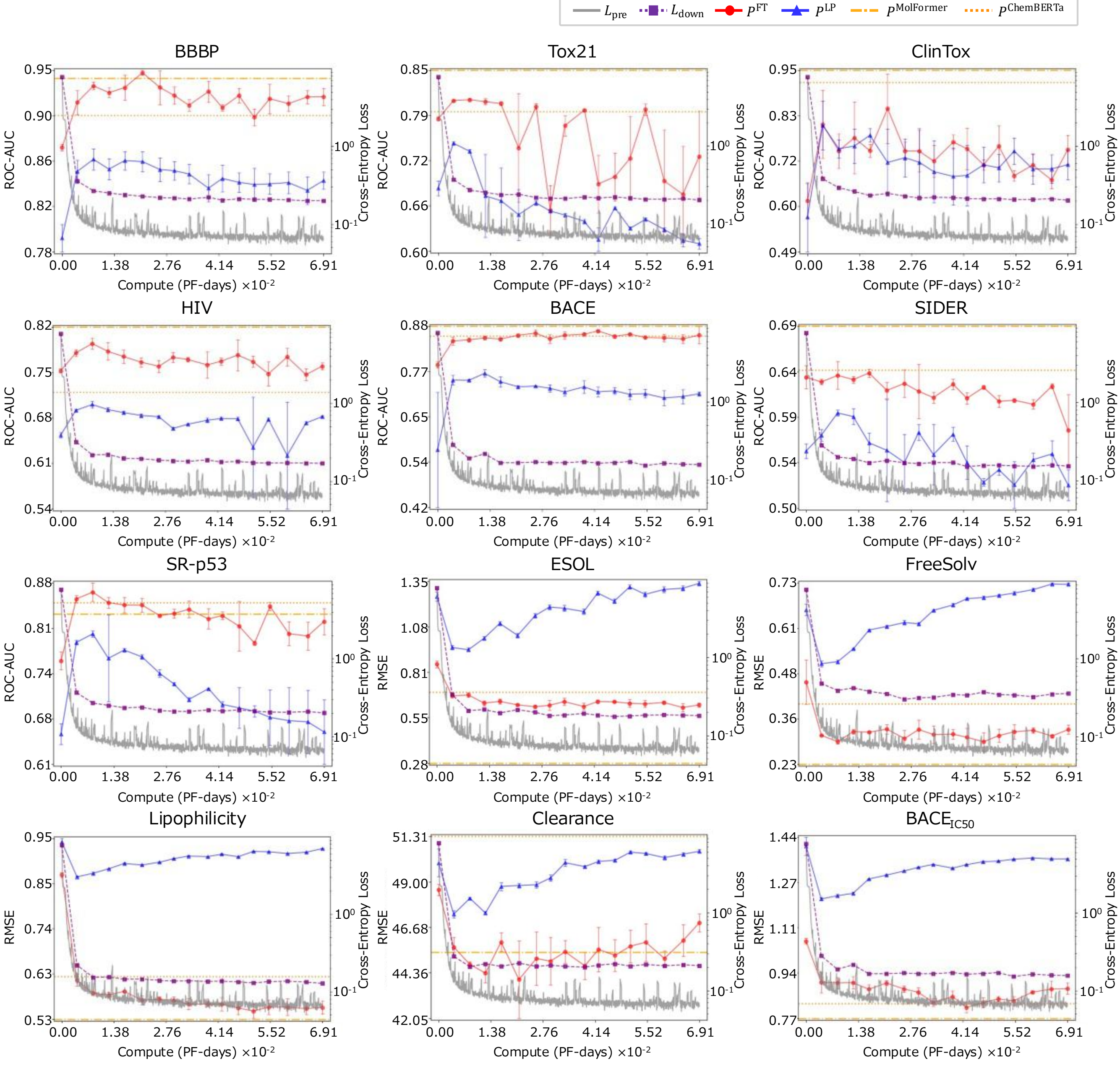

Figure 15: Task-wise curves under compute scaling. The x-axis is the pre-training compute measured in PF-days. Other axes and plotted quantities are the same as Figure 9.

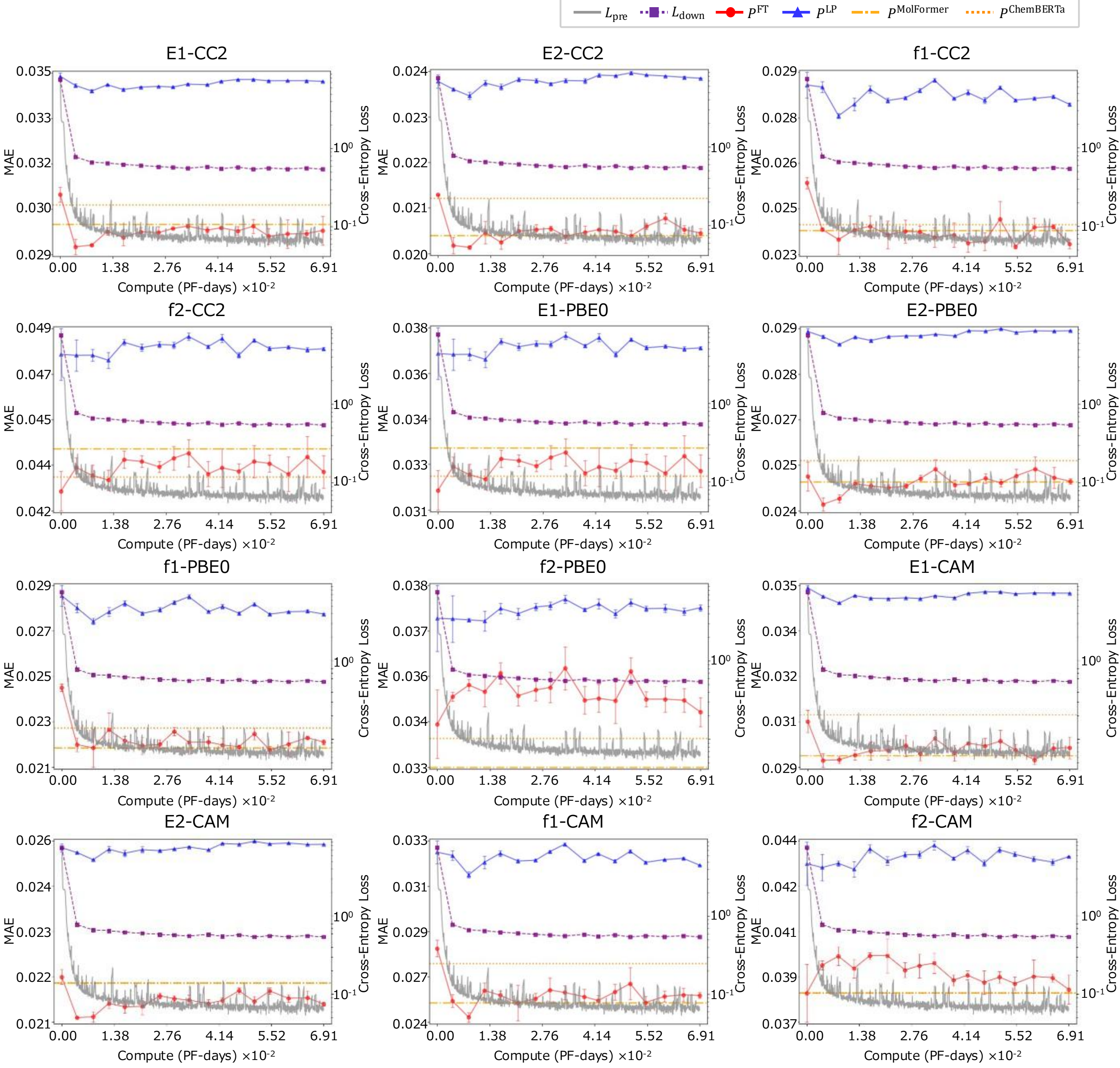

Figure 16: Task-wise curves under compute scaling for QM8 tasks. Axes and plotted quantities follow Figure 15.

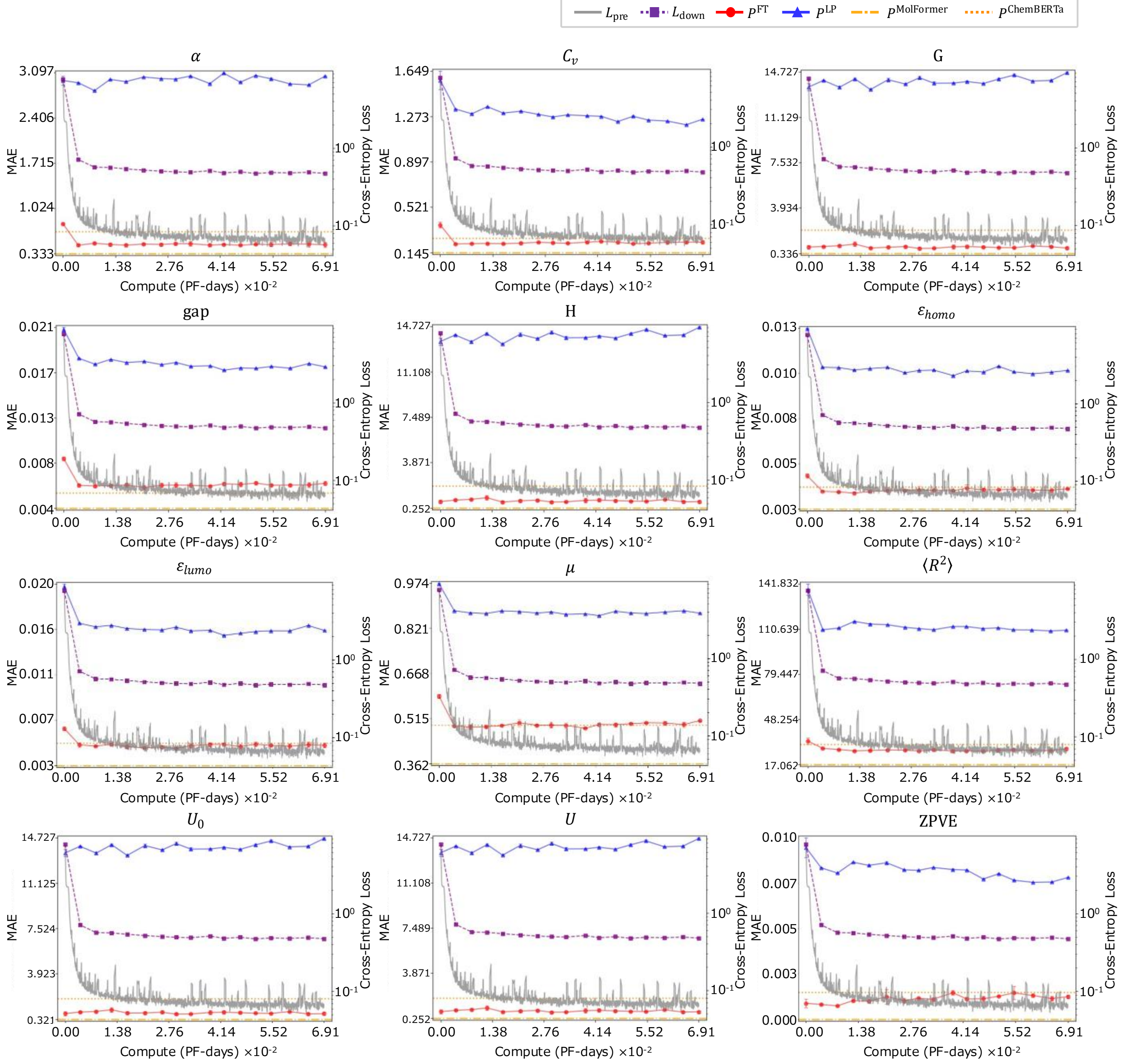

Figure 17: Task-wise curves under compute scaling for QM9 tasks. Axes and plotted quantities follow Figure 15.

## H Dataset Statistics

This appendix summarizes two types of information to complement the task-wise correlations between pre-training loss $L_{\mathrm{pre}}$ and downstream performance reported in Figure 4. First, we provide a per-task list of Spearman's rank correlation coefficients $\rho$ in Table 10. Second, to examine whether the cross-task variation in these correlations can be explained by simple dataset-level factors, we report dataset statistics for each benchmark in Table 11 and 12.

Table 10 reports per-task values of $\rho(L_{\mathrm{pre}}, P^{\mathrm{FT}})$ and $\rho(L_{\mathrm{pre}}, P^{\mathrm{LP}})$ computed over checkpoint sequences obtained under settings where only one training resource is scaled at a time, namely model size, data size, or compute.

Table 11 summarizes statistics of inputs and targets for each benchmark. Input statistics include the number of samples, token length, vocabulary size, and out-of-vocabulary (OOV) rate. Target statistics include class imbalance for classification tasks, and skewness and kurtosis of the target distribution for regression tasks. Table 12 reports molecular-property statistics including the mean molecular weight, mean heavy-atom count, mean ring count, and mean LogP.

Qualitative comparison of the statistics in Table 11 and 12 with the correlations in Table 10 did not reveal a clear, consistent trend explaining the cross-task variation in correlation coefficients within the ranges considered in this study. This suggests that the task-dependent relationship between pre-training loss and downstream performance may not be fully captured by such simple dataset statistics alone

| Benchmark | $\rho(L_{\text{pre}}, P^{\text{FT}})$ | | | $\rho(L_{\text{pre}}, P^{\text{LP}})$ | | |
|---|---|---|---|---|---|---|
| | Model | Data | Compute | Model | Data | Compute |
| ESOL | 1.00 | 0.40 | -0.72 | 1.00 | 1.00 | 0.45 |
| FreeSolv | 1.00 | 0.40 | -0.72 | 0.89 | 0.70 | -0.03 |
| Lipophilicity | 0.96 | 0.40 | -0.48 | 0.96 | 1.00 | 0.75 |
| HIV | 1.00 | 0.80 | -0.46 | 0.96 | -0.10 | -0.25 |
| BACE | 0.82 | -0.40 | -0.52 | 0.75 | 0.90 | 0.42 |
| $\text{BACE}_{\text{IC50}}$ | 1.00 | -0.90 | -0.55 | 1.00 | 0.90 | 0.62 |
| BBBP | 0.75 | 0.60 | -0.25 | 0.96 | 0.60 | -0.20 |
| Tox21 | 1.00 | -0.30 | -0.77 | 0.96 | 0.40 | -0.74 |
| SR-p53 | 0.96 | -0.60 | -0.52 | 0.96 | 0.70 | -0.39 |
| SIDER | 0.89 | -0.20 | -0.62 | 0.96 | 0.10 | -0.86 |
| ClinTox | 0.96 | 0.80 | -0.25 | 0.89 | -0.60 | -0.33 |
| Clearance | 1.00 | -0.70 | -0.74 | 0.89 | 0.60 | -0.13 |
| E1-CC2 | 0.36 | 0.90 | -0.32 | 0.96 | 0.60 | -0.12 |
| E2-CC2 | 0.39 | 0.60 | -0.66 | 0.86 | -0.90 | -0.40 |
| f1-CC2 | 0.96 | 0.70 | 0.21 | -0.07 | 1.00 | 0.43 |
| f2-CC2 | 0.96 | -0.10 | -0.35 | -0.11 | 0.90 | -0.27 |
| E1-PBE0 | 0.32 | 0.80 | -0.20 | 0.96 | 0.60 | -0.31 |
| E2-PBE0 | 0.39 | 0.70 | -0.63 | 0.86 | -0.20 | -0.54 |
| f1-PBE0 | 0.96 | 0.90 | 0.20 | 0.11 | 0.10 | 0.20 |
| f2-PBE0 | 0.89 | -0.60 | -0.61 | -0.18 | 0.10 | 0.18 |
| E1-CAM | 0.39 | 0.80 | -0.27 | 0.96 | 0.60 | -0.09 |
| E2-CAM | 0.32 | 0.30 | -0.71 | 0.86 | 0.40 | -0.36 |
| f1-CAM | 0.96 | 0.90 | 0.05 | -0.07 | 0.60 | -0.01 |
| f2-CAM | 0.68 | -0.50 | -0.46 | 0.21 | 0.70 | 0.32 |
| $\alpha$ | 1.00 | 0.10 | -0.26 | 0.89 | 1.00 | -0.02 |
| $C_v$ | 1.00 | 0.90 | 0.87 | 0.96 | 0.70 | -0.34 |
| G | 0.96 | 0.10 | -0.47 | 0.89 | 0.70 | 0.34 |
| Gap | 1.00 | 1.00 | 0.71 | 0.96 | 0.60 | -0.35 |
| H | 0.96 | 0.10 | -0.47 | 0.89 | 0.70 | 0.17 |
| $\varepsilon_{homo}$ | 1.00 | 1.00 | 0.56 | 1.00 | 0.70 | -0.56 |
| $\varepsilon_{lumo}$ | 1.00 | 1.00 | 0.68 | 1.00 | 0.70 | 0.31 |
| $\mu$ | 1.00 | 1.00 | 0.32 | 0.96 | 0.70 | -0.45 |
| $\langle R^2 \rangle$ | 1.00 | 0.90 | 0.66 | 1.00 | 0.60 | 0.29 |
| $U_0$ | 1.00 | 0.10 | -0.47 | 0.89 | 0.70 | 0.34 |
| $U$ | 0.96 | 0.10 | -0.47 | 0.89 | 0.60 | 0.29 |
| ZPVE | 1.00 | 0.90 | 0.77 | 0.96 | -0.70 | -0.67 |

Table 10: Per-task Spearman's rank correlation coefficients $\rho$. For each checkpoint sequence where only one factor is scaled (model size, data size, or compute), we compute $\rho(L_{\text{pre}}, P^{\text{FT}})$ and $\rho(L_{\text{pre}}, P^{\text{LP}})$.

| Benchmark | Input statistics | | | | Target Statistics | | |
|---|---|---|---|---|---|---|---|
| | # samples | Token length (min/mean/max) | Vocab size | OOV rate | Class imbalance ratio | Skewness | Kurtosis |
| ESOL | 1127 | (1, 21.49, 97) | 29 | 0.00 | - | -0.49 | 3.18 |
| FreeSolv | 641 | (1, 13.16, 44) | 26 | 0.00 | - | -1.17 | 6.62 |
| Lipophilicity | 4200 | (10, 44.83, 205) | 33 | 0.00 | - | -0.58 | 2.88 |
| HIV | 41126 | (2, 43.43, 404) | 188 | 0.00 | 27.5 | - | - |
| BACE | 1512 | (14, 58.82, 174) | 30 | 0.00 | 1.19 | - | - |
| $BACE_{IC50}$ | 1512 | (14, 58.82, 174) | 30 | 0.00 | - | -0.49 | 2.63 |
| BBBP | 2038 | (3, 40.80, 234) | 51 | 0.00 | 3.25 | - | - |
| Tox21 | 7830 | (1, 30.54, 240) | 132 | 0.00 | 16.9 | - | - |
| SR-p53 | 6773 | (1, 29.60, 240) | 127 | 0.00 | 15.0 | - | - |
| SIDER | 1426 | (1, 58.27, 897) | 103 | 0.00 | 6.40 | - | - |
| ClinTox | 1476 | (1, 44.74, 242) | 69 | 0.00 | 13.4 | - | - |
| Clearance | 837 | (17, 48.47, 141) | 30 | 0.00 | - | -0.11 | 1.68 |
| E1-CC2 | 21747 | (1, 12.63, 20) | 31 | 0.00 | - | -0.13 | 2.90 |
| E2-CC2 | 21747 | (1, 12.63, 20) | 31 | 0.00 | - | -0.26 | 3.62 |
| f1-CC2 | 21747 | (1, 12.63, 20) | 31 | 0.00 | - | 3.69 | 19.28 |
| f2-CC2 | 21747 | (1, 12.63, 20) | 31 | 0.00 | - | 2.50 | 9.60 |
| E1-PBE0 | 21747 | (1, 12.63, 20) | 31 | 0.00 | - | -0.04 | 2.86 |
| E2-PBE0 | 21747 | (1, 12.63, 20) | 31 | 0.00 | - | -0.13 | 3.16 |
| f1-PBE0 | 21747 | (1, 12.63, 20) | 31 | 0.00 | - | 3.31 | 15.17 |
| f2-PBE0 | 21747 | (1, 12.63, 20) | 31 | 0.00 | - | 2.77 | 11.44 |
| E1-CAM | 21747 | (1, 12.63, 20) | 31 | 0.00 | - | -0.07 | 2.71 |
| E2-CAM | 21747 | (1, 12.63, 20) | 31 | 0.00 | - | -0.17 | 3.01 |
| f1-CAM | 21747 | (1, 12.63, 20) | 31 | 0.00 | - | 3.19 | 14.06 |
| f2-CAM | 21747 | (1, 12.63, 20) | 31 | 0.00 | - | 2.49 | 9.46 |
| $\alpha$ | 133885 | (1, 14.78, 22) | 30 | 0.00 | - | -0.38 | 4.80 |
| $C_v$ | 133885 | (1, 14.78, 22) | 30 | 0.00 | - | -0.06 | 3.42 |
| G | 133885 | (1, 14.78, 22) | 30 | 0.00 | - | 0.32 | 4.94 |
| Gap | 133885 | (1, 14.78, 22) | 30 | 0.00 | - | -0.01 | 2.44 |
| H | 133885 | (1, 14.78, 22) | 30 | 0.00 | - | 0.32 | 4.94 |
| $\varepsilon_{homo}$ | 133885 | (1, 14.78, 22) | 30 | 0.00 | - | 0.27 | 4.71 |
| $\varepsilon_{lumo}$ | 133885 | (1, 14.78, 22) | 30 | 0.00 | - | -0.18 | 2.41 |
| $\mu$ | 133885 | (1, 14.78, 22) | 30 | 0.00 | - | 1.46 | 11.36 |
| $\langle R^2 \rangle$ | 133885 | (1, 14.78, 22) | 30 | 0.00 | - | 1.29 | 6.86 |
| $U_0$ | 133885 | (1, 14.78, 22) | 30 | 0.00 | - | 0.32 | 4.94 |
| $U$ | 133885 | (1, 14.78, 22) | 30 | 0.00 | - | 0.32 | 4.94 |
| ZPVE | 133885 | (1, 14.78, 22) | 30 | 0.00 | - | 0.01 | 3.00 |

Table 11: Input and target statistics for each benchmark. We report the number of samples, token length, vocabulary size, and out-of-vocabulary (OOV) rate. We also report target statistics, including class imbalance for classification tasks, as well as skewness and kurtosis of the target distribution for regression tasks.

| Benchmark | Chemical statistics | | | |
| --- | --- | --- | --- | --- |
| | Molecule weight | Heavy atom count | Ring count | LogP |
| ESOL | 203.79 | 13.28 | 1.39 | 2.44 |
| FreeSolv | 139.03 | 8.73 | 0.66 | 1.94 |
| Lipophilicity | 383.14 | 27.04 | 3.49 | 3.28 |
| HIV | 370.10 | 25.51 | 3.07 | 2.98 |
| BACE | 479.79 | 34.10 | 3.78 | 3.13 |
| $BACE_{IC50}$ | 479.79 | 34.10 | 3.78 | 3.13 |
| BBBP | 344.79 | 24.06 | 2.99 | 2.32 |
| Tox21 | 276.14 | 18.57 | 1.77 | 2.37 |
| SR-p53 | 266.12 | 17.90 | 1.69 | 2.23 |
| SIDER | 493.06 | 33.64 | 3.00 | 1.26 |
| ClinTox | 383.55 | 26.19 | 2.76 | 1.30 |
| Clearance | 413.98 | 29.08 | 3.59 | 3.49 |
| E1-CC2 | 108.86 | 7.77 | 1.41 | 0.30 |
| E2-CC2 | 108.86 | 7.77 | 1.41 | 0.30 |
| f1-CC2 | 108.86 | 7.77 | 1.41 | 0.30 |
| f2-CC2 | 108.86 | 7.77 | 1.41 | 0.30 |
| E1-PBE0 | 108.86 | 7.77 | 1.41 | 0.30 |
| E2-PBE0 | 108.86 | 7.77 | 1.41 | 0.30 |
| f1-PBE0 | 108.86 | 7.77 | 1.41 | 0.30 |
| f2-PBE0 | 108.86 | 7.77 | 1.41 | 0.30 |
| E1-CAM | 108.86 | 7.77 | 1.41 | 0.30 |
| E2-CAM | 108.86 | 7.77 | 1.41 | 0.30 |
| f1-CAM | 108.86 | 7.77 | 1.41 | 0.30 |
| f2-CAM | 108.86 | 7.77 | 1.41 | 0.30 |
| $\alpha$ | 122.76 | 8.80 | 1.74 | 0.30 |
| $C_v$ | 122.76 | 8.80 | 1.74 | 0.30 |
| G | 122.76 | 8.80 | 1.74 | 0.30 |
| Gap | 122.76 | 8.80 | 1.74 | 0.30 |
| H | 122.76 | 8.80 | 1.74 | 0.30 |
| $\varepsilon_{homo}$ | 122.76 | 8.80 | 1.74 | 0.30 |
| $\varepsilon_{lumo}$ | 122.76 | 8.80 | 1.74 | 0.30 |
| $\mu$ | 122.76 | 8.80 | 1.74 | 0.30 |
| $\langle R^2 \rangle$ | 122.76 | 8.80 | 1.74 | 0.30 |
| $U_0$ | 122.76 | 8.80 | 1.74 | 0.30 |
| $U$ | 122.76 | 8.80 | 1.74 | 0.30 |
| ZPVE | 122.76 | 8.80 | 1.74 | 0.30 |

Table 12: Chemical statistics of molecules for each benchmark. We report the mean molecular weight, mean heavy-atom count, mean ring count, and mean LogP.

# I Task-wise Trajectories of Metrics Beyond Pre-training Loss

Following the protocol in Section 3.3, this appendix provides task-wise visualizations that complement the aggregated results reported in Section 4.3. We use the 4.87M-parameter model pre-trained on 42.4B tokens, and compute $\mathbf{Tr}(\boldsymbol{H})$ and the PGM distance at checkpoints saved every 0.25 epoch during pre-training as described in Appendix C. For each downstream task, we plot these metric trajectories alongside downstream performance under fine-tuning and linear probe, where downstream results are averaged over three random seeds. Figure 18–20 report the per-task results for classification, non-quantum regression, and quantum-chemistry regression tasks, respectively. Together, these plots illustrate how loss-alternative metrics evolve over training at the per-task level and how their trajectories relate to downstream outcomes. For some combinations of checkpoints and tasks, the computation of the PGM distance was numerically unstable, resulting in missing values, and these cases are therefore not visualized in Figure 18–20. In Section 4.3, average metric values for each task group were computed after excluding these missing entries.

# J Task-wise Parameter Visualization

This appendix extends the parameter-space analysis in Section 4.4 by presenting the visualization separately for each downstream task. We use the same pre-training run as the main text and consider the sequence of checkpoints saved every 0.25 epoch during pre-training together with corresponding task-specific fine-tuned models. For each pre-training checkpoint and each fine-tuned model, we extract the parameter vector of the final Transformer block, and project all models into two dimensions using PCA. Importantly, PCA is fit once on the union of all pre-training checkpoints and all fine-tuned models across tasks, and the same projection is used for every task-wise plot, enabling consistent comparisons across tasks. To highlight changes induced by fine-tuning relative to each initialization, we report the visualization in relative coordinates: for each task and each initialization checkpoint, we translate the projected coordinates so that the initialization checkpoint is placed at the origin. Figures 21–23 report these task-wise visualizations, grouped in the same way as earlier sections: classification and non-quantum regression tasks (Figure 21), QM8 tasks (Figure 22), and QM9 tasks (Figure 23).

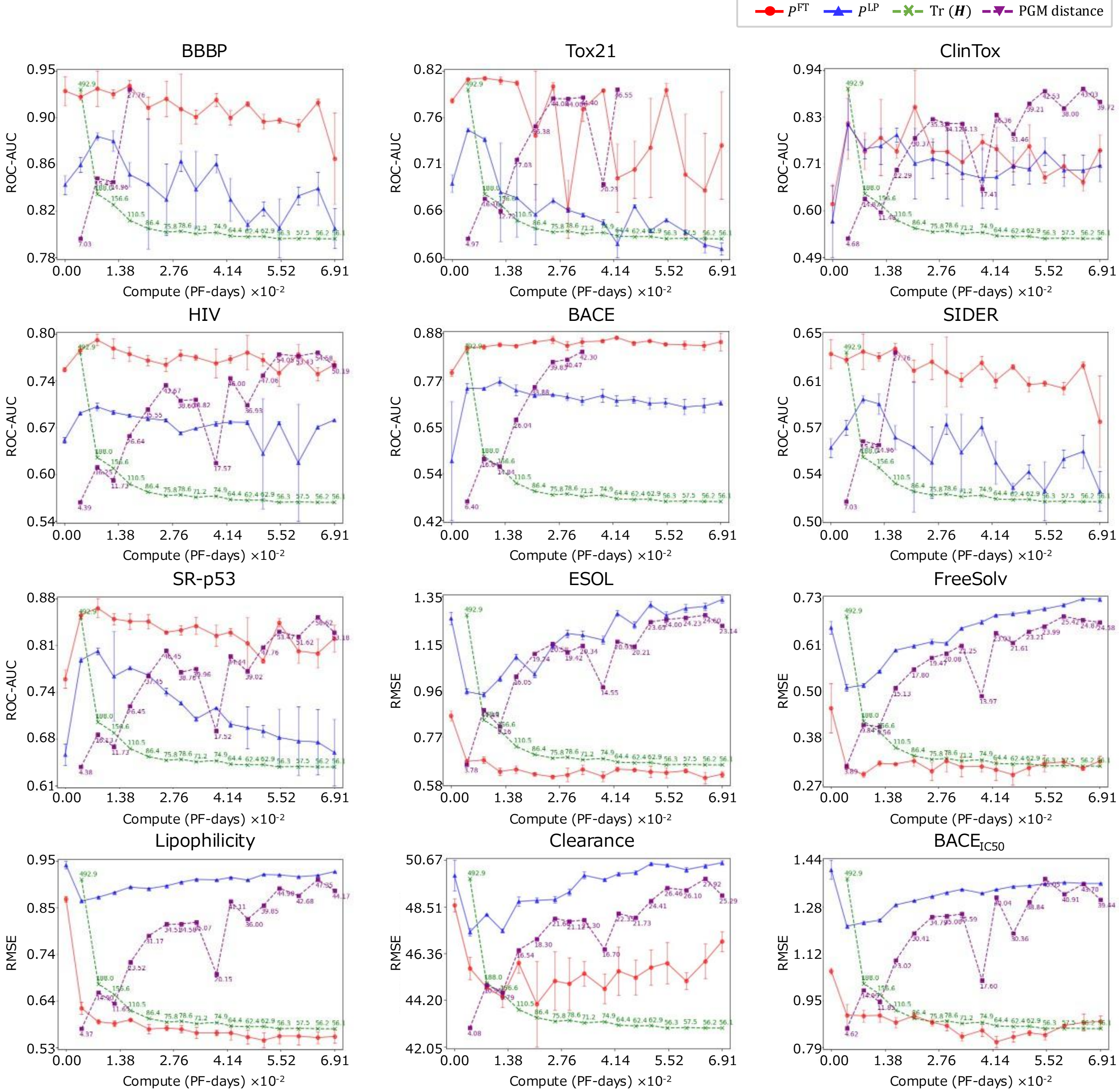

Figure 18: Trajectories of metrics beyond pre-training loss for classification and non-quantum regression tasks.

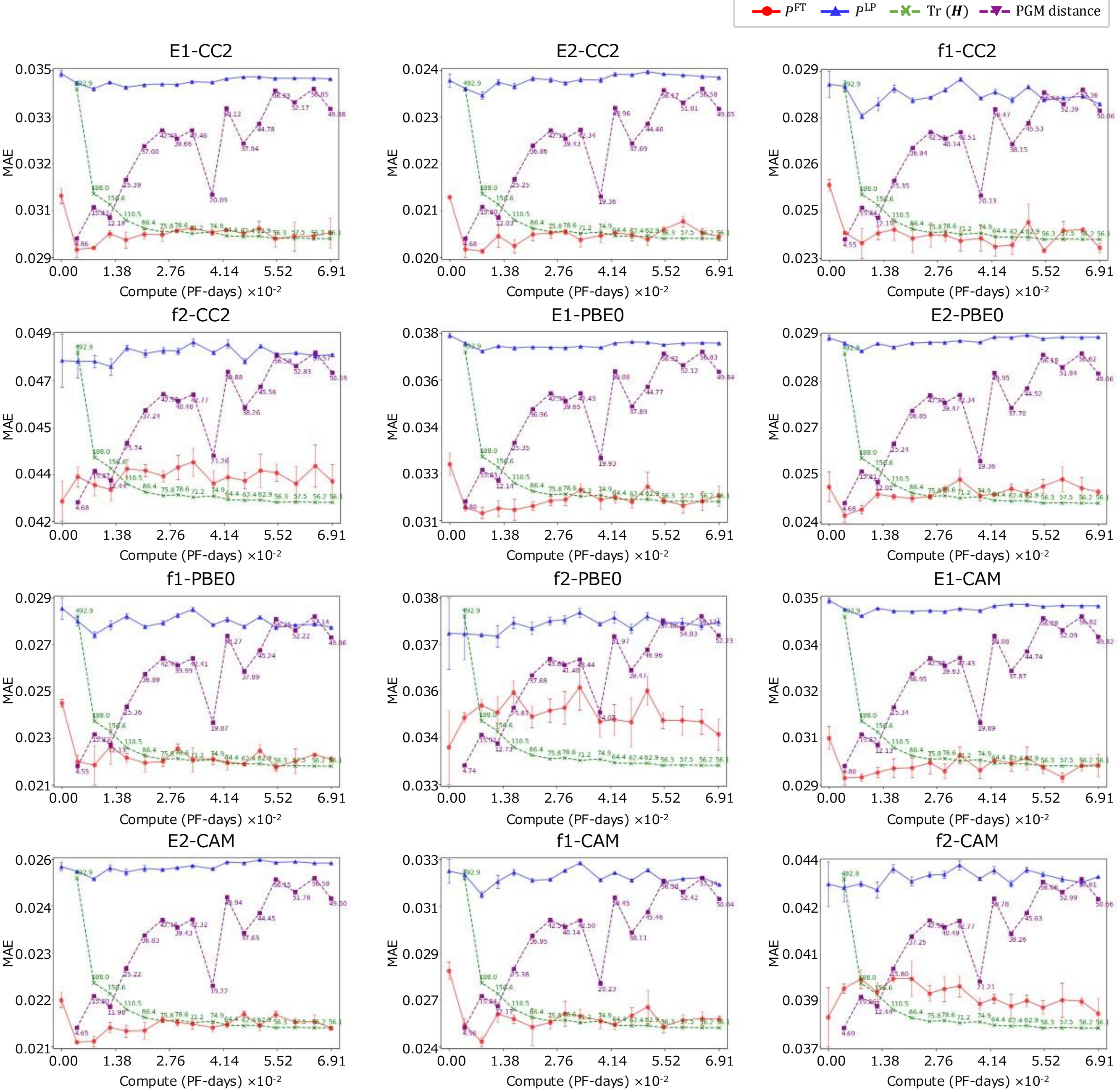

Figure 19: Trajectories of loss-alternative metrics for QM8 tasks.

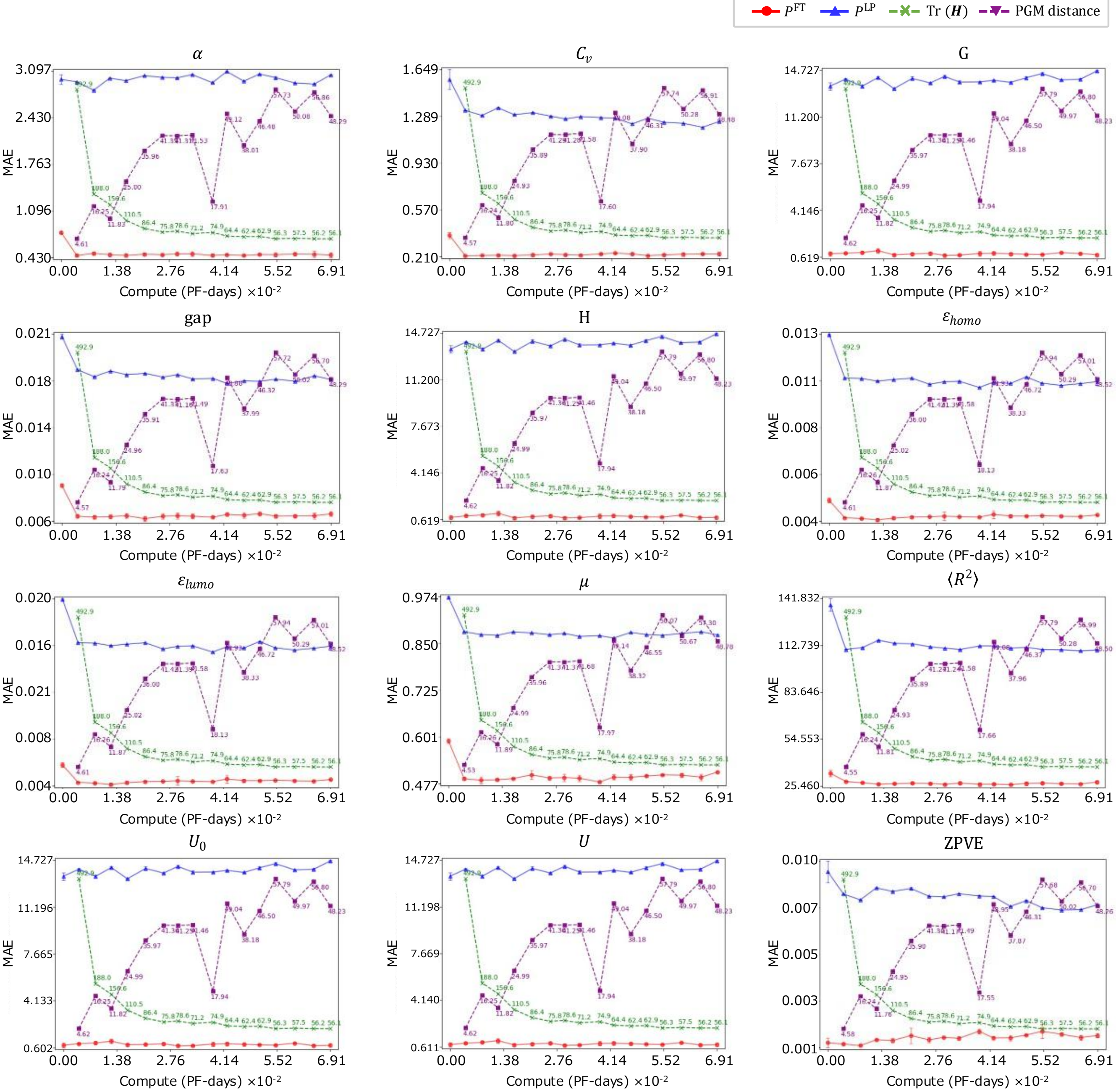

Figure 20: Trajectories of loss-alternative metrics for QM9 tasks.

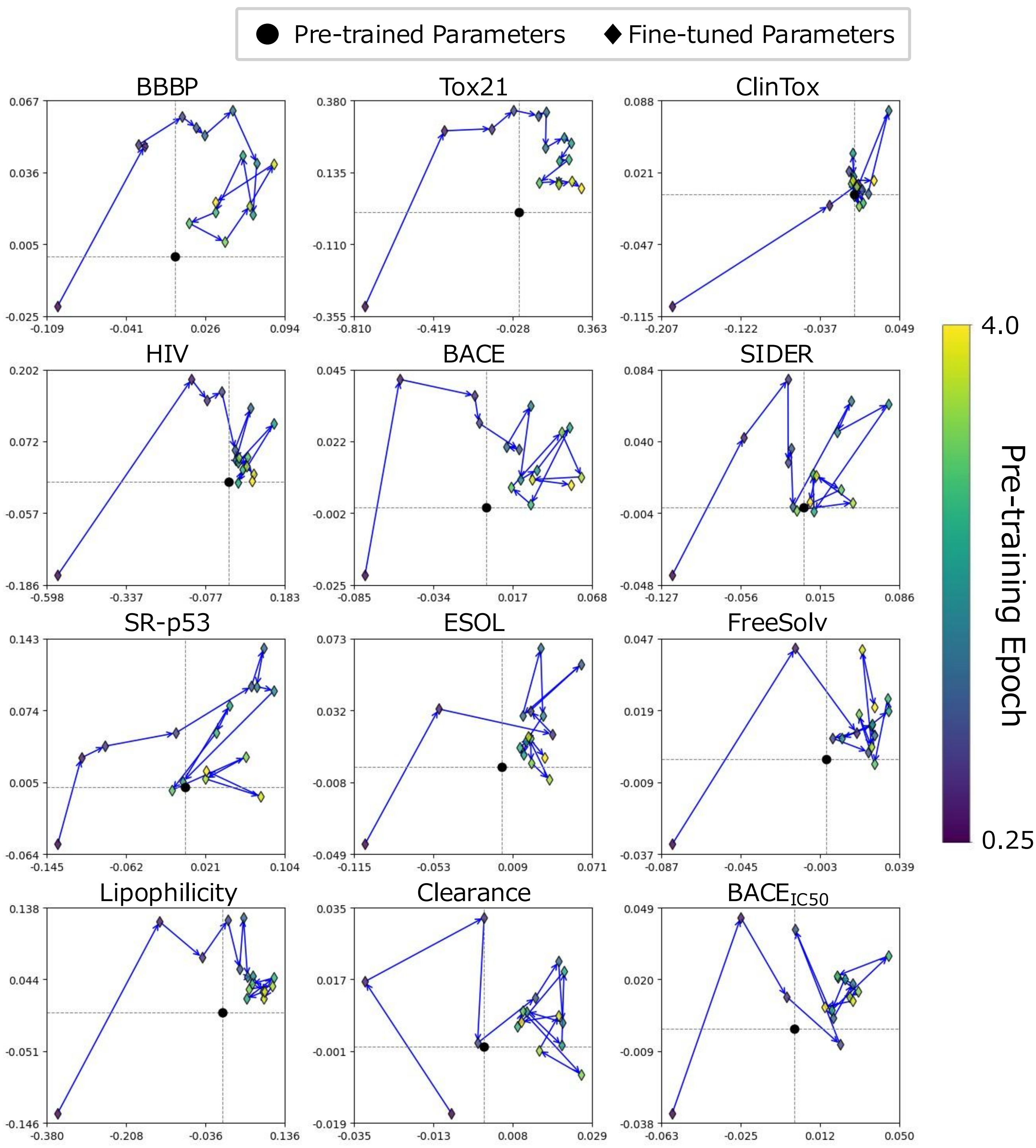

Figure 21: Task-wise parameter-space visualizations for classification and non-quantum regression tasks.

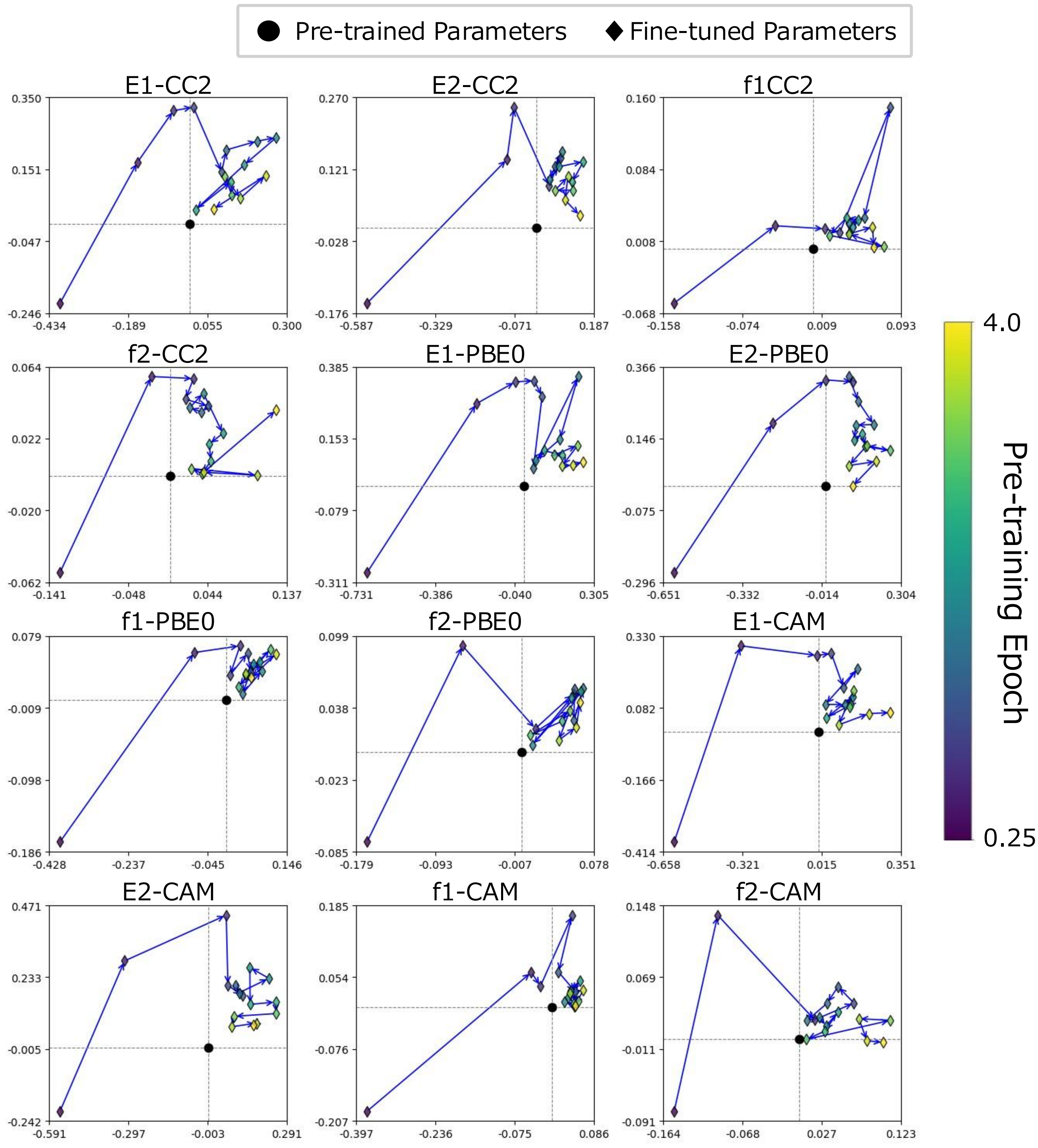

Figure 22: Task-wise parameter-space visualizations for QM8 tasks.

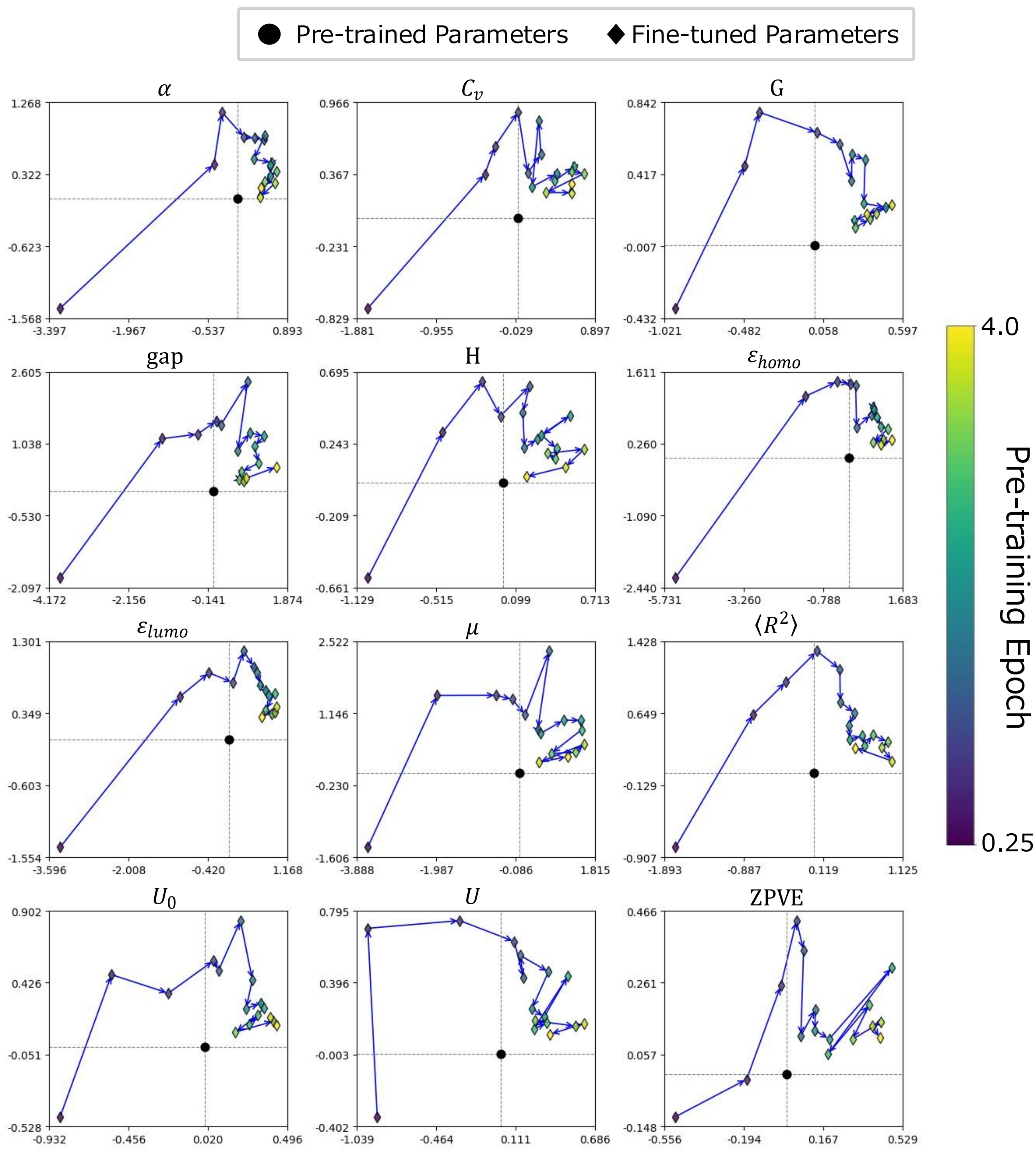

Figure 23: Task-wise parameter-space visualizations for QM9 tasks.